\documentclass{article}

\usepackage[preprint]{neurips_2024}

\usepackage[utf8]{inputenc} 
\usepackage[T1]{fontenc}    
\usepackage{hyperref}       
\usepackage{url}            
\usepackage{booktabs}       
\usepackage{amsfonts}       
\usepackage{nicefrac}       
\usepackage{microtype}      
\usepackage{xcolor}  %
\usepackage{amsmath}
\usepackage{amssymb}
\usepackage{mathtools}
\usepackage{amsthm}
\usepackage{multirow}
\usepackage{makecell}
\usepackage{tabularx}
\usepackage{xspace}
\usepackage{wrapfig}

\usepackage{bm}
\usepackage{diagbox}
\usepackage{courier}
\usepackage{subcaption}
\usepackage{natbib}
\setcitestyle{numbers,square}

\definecolor{mypurple}{RGB}{126, 116, 212}
\def\eg{\emph{e.g.}\xspace}
\def\ie{\emph{i.e.}\xspace}
\def\wrt{\emph{w.r.t.}\xspace}

\def\fontchange#1#2{{\fontfamily{#1}\selectfont#2}}

\definecolor{dpurple}{RGB}{128, 0, 128}
\definecolor{lpurple}{RGB}{235, 232, 242}

\title{Where is the answer? Investigating Positional Bias in Language Model Knowledge Extraction}

\author{%
  Kuniaki Saito\thanks{OMRON SINIC X, kuniaki.saito@sinicx.com. All experiments are conducted by OMRON SINIC X.} \\    
  \\\\
   \And
   Kihyuk Sohn\thanks{Google Research} \\
  \And
  Chen-Yu Lee\thanks{Google Cloud AI Research}\\
  \And 
  Yoshitaka Ushiku\textsuperscript{*}
}

\begin{document}

\maketitle

\begin{abstract}
Large language models require updates to remain up-to-date or adapt to new domains by fine-tuning them with new documents. One key is memorizing the latest information in a way that the memorized information is \textit{extractable} with a query prompt. However, LLMs suffer from a phenomenon called “perplexity curse”; despite minimizing document perplexity during fine-tuning, LLMs struggle to extract information through a prompt sentence. 
In this new knowledge acquisition and extraction, we find a very intriguing fact that LLMs can accurately answer questions about the first sentence, but they struggle to extract information described in the middle or end of the documents used for fine-tuning. Our study suggests that the auto-regressive training causes this issue; each token is prompted by reliance on all previous tokens, which hinders the model from recalling information from training documents by question prompts. To conduct the in-depth study, we publish both synthetic and real datasets, enabling the evaluation of the QA performance \wrt the position of the corresponding answer in a document. Our investigation shows that even a large model suffers from the “perplexity curse”, but regularization such as denoising auto-regressive loss can enhance the information extraction from diverse positions. These findings will be (i) a key to improving knowledge extraction from LLMs and (ii) new elements to discuss the trade-off between RAG and fine-tuning in adapting LLMs to a new domain.


\end{abstract}

\section{Introduction}
\label{intro}
Large language model stores diverse factual knowledge in their parameters, which is learned during self-supervised training on a huge amount of document data and is made extractable in the form of question answering by instruction-tuning~\cite{touvron2023llama,workshop2022bloom,touvron2023llama2,penedo2023refinedweb,brown2020language,raffel2020exploring,gao2020pile,wei2021finetuned}. Since the knowledge of LLM is limited to the training data publicly available in a certain period, such as Wikipedia, Github, and CommonCrawl, the models, of course, cannot retrieve information about unseen domains. Therefore, efficiently tailoring LLM to a specific field or updating the factual knowledge to new information is very important~\cite{han2023medalpaca,wu2023pmc,li2023llava}. To keep LLM's knowledge up-to-date or adapt it to a new domain, tuning it on new documents is essential~\cite{jang2022towards,hu2023meta}. 
This fine-tuning requires training corpora, \ie, a collection of new documents, and question-answering (QA) data asking about the factual knowledge in the documents~\cite{allen2023physics,jiang2024instruction}. The documents are the source of new knowledge while the QA data should enhance the extraction of diverse contents~\cite{zhu2023physics}. 
One key is memorizing the new fact so that the memorized information is \textit{extractable} with a query prompt. The standard fine-tuning paradigm tunes the models with an auto-regressive objective; the model is trained to predict the next token given previous tokens, yet it is unclear whether such naive training is optimal to store the knowledge in an \textit{extractable} manner. In fact, \cite{jiang2024instruction} report a phenomenon called “perplexity curse”; the amount of elicited knowledge is limited even though the perplexity of documents is minimized.

\begin{figure}
    \centering
\includegraphics[width=\textwidth]{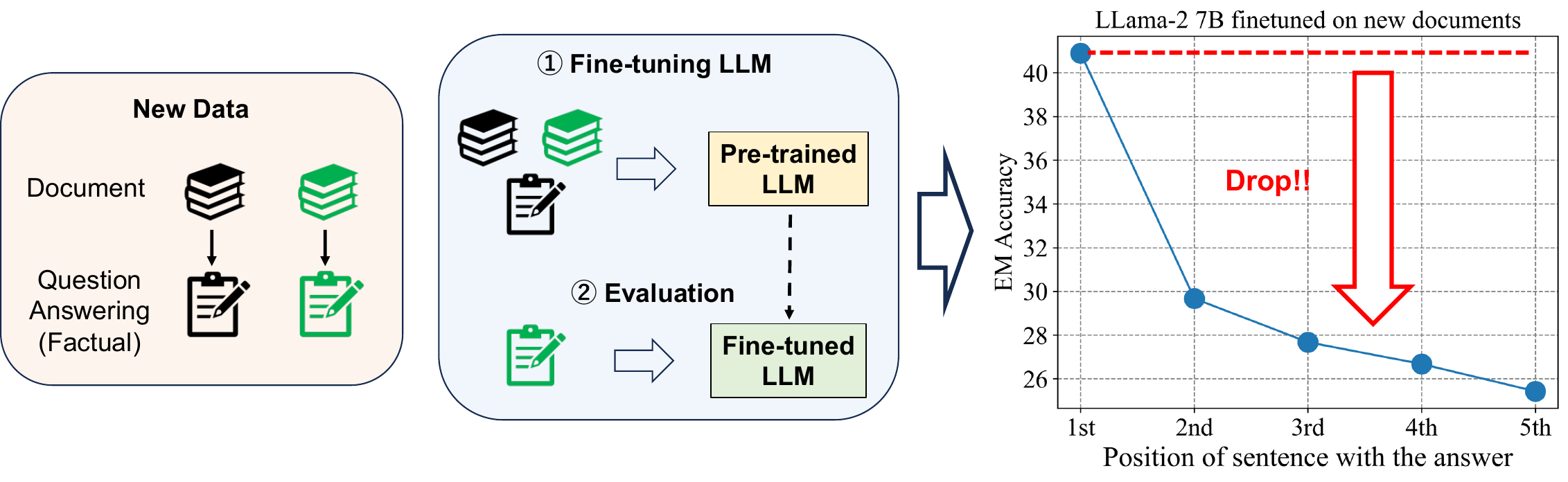}
\vspace{-3mm}
    \caption{\textbf{Left}: In our problem setting, we fine-tune a pre-trained LLM on new documents and QA data created from the documents, and evaluate on hold-out QA data (green). \textbf{Right}: We vary the position of the answer sentence in training documents and fine-tune models on the modulated documents. Each plot shows a model tuned on documents with different answer positions. We find that the fine-tuned LLM suffers from the positional bias issue, \ie, it cannot accurately answer questions described in the middle or the end of documents. See Sec.~\ref{sec:answer_position} for details.}
    \label{fig:neurips_teaser}
\end{figure}
To conduct an in-depth study of the “perplexity curse”, we employ synthetic and real document datasets, finding a very intriguing fact that fine-tuned LLM suffers from positional bias in the training corpora, \ie, it struggles to answer questions about the information described in the middle or at the end of the training document as shown in Fig.~\ref{fig:neurips_teaser}. 
This phenomenon differs from the reported “Lost in the middle” behavior~\cite{liu2023lost}, which reports that LLMs struggle to extract information from the middle of the context passage given during inference. Instead, we focus on the training stage, therefore, the documents are not present in the prompt, but in training corpora. 
Our empirical results suggest that the vanilla auto-regressive model memorizes the learned documents well but in a hard-to-extract way. 
Specifically, an analysis of the perplexity indicates that the model reconstructs each sentence by relying much on previous tokens. This prevents the model from extracting factual knowledge given a query sentence that differs from the tokens prompting the knowledge in documents. To minimize the perplexity of documents during training, the model seems to learn the spurious correlation between tokens to prompt the next token, which hinders memorizing information in an extractable manner. 

Our analysis demonstrates that several existing techniques can mitigate the issue, yet are overlooked. In particular, we discover that denoising auto-regressive training, which randomly replaces input tokens with different ones, can significantly improve the ability to access information from the middle or end of the learned documents. The technique is simple and easy to plug in, yet surprisingly effective to address the issue. 
Note that we do not aim to claim the novelty of this technique. Instead, we emphasize the potential risks of relying solely on the vanilla auto-regressive objective, despite its prevalence in current practice.
Also, our insight will be new ingredients to discuss the trade-off between retrieval augmented generation and fine-tuning-based approaches to answer factual knowledge in a new domain. 

Our contributions are summarized as follows: 
\begin{itemize}
    \item We discover that LLM suffers from positional bias within documents, \ie, LLM struggles to extract information described in the middle or the end of the training document when applied to a new domain.    
    \item To encourage further development to solve the problem, we publish a synthetic dataset and documents that are collected from Wikipedia. They contain QA pairs and annotations to the position of the sentence to which the question refers. 
    \item We provide an in-depth study of these datasets and show that several regularization techniques such as denoising auto-regressive training can lead to remarkable improvement in synthetic, Wikipedia, and medical documents datasets. This finding will be useful for future research in adapting LLM to a new domain via fine-tuning.
\end{itemize}

\vspace{-3mm}
\section{Related Work}
\label{related}
\vspace{-3mm}
\textbf{Ability to memorize factual knowledge.}
\cite{zhu2023physics} analyze whether LM answers questions based on exposure to exact/similar questions during instruction-tuning, or if it extracts knowledge learned in pre-training. They show that LM can memorize and extract information embedded in the training corpora even though the model does not see the exact question during training. Therefore, how to learn from the training corpora should be a key factor. 

\textbf{Continual adaptation of LLM.}
\cite{jang2022towards, hu2023meta} explore continually adapting a small language model to a new domain. \cite{hu2023meta} introduce a meta-learning approach for adaptation, requiring a lot of computation, thus not suitable to fine-tune LLM. Our studied recipes incur a negligible increase in training costs yet bring significant improvement in QA performance. \cite{jiang2024instruction} introduces \textit{pre-instruction-tuning}, a method that instruction-tunes model on questions before training on documents. Their approach involves additional hyper-parameters \eg, training epochs and learning rate in each stage. Notably, due to its simplicity, our recipes should be easy to plug into their method\footnote{We could not reproduce their results probably because of hyper-parameters. Combining with their approach is our future work.}. Also, tailoring many open-source LLMs for a specific domain is popular, \eg, medical domain~\cite{han2023medalpaca,wu2023pmc,li2023llava}, which is the important application of continual adaptation of LLM. Our results on the medical dataset suggest that regularization enhances the memorization of knowledge in an extractable way.
Thus, our argument about the positional bias in LLM should benefit diverse communities developing customized LLMs. 

\textbf{Overwriting factual knowledge.} 
Many approaches have been proposed to edit the knowledge in LM~\cite{mitchell2021fast,mitchell2022memory,meng2022locating,meng2022memit,feigenbaum2024editing,wang2023knowledge}. Their interest is not in \textit{adding} new knowledge, but in \textit{editing} existing knowledge, \eg, “Donald Trump is the President of USA”, into a new one, “Joe Biden is the President of USA”. Also, they handle facts represented as a single sentence while we focus on how to store and retrieve new knowledge, often represented by 5-10 sentences. 

\textbf{Retrieval augmented generation (RAG).}
One alternative approach to answering questions about new documents is to use RAG~\cite{lewis2020retrieval,guu2020retrieval,hofstatter2022multi} with LLMs, \ie, retrieving several documents and deriving answers based on them. While our results indicate that RAG can show high performance in answering factual knowledge, it requires a powerful retrieval model to efficiently search through documents before prediction, and the LLM needs to be able to handle the resulting long context. As shown in the discussion of the pros and cons of RAG and fine-tuning approaches in Sec.~\ref{sec:wiki2023}, we do not argue that fine-tuning surpasses RAG in adapting to a new domain, instead, we should choose a better framework depending on the applications or unifying two frameworks is an interesting direction. 

\textbf{Positional bias in LLM.}
It is widely known that LLM suffers from the so-called \textit{positional bias} issue~\cite{liu2023lost, ko2020look,ma2021exploiting,hofstatter2021mitigating,glater2023answerpos}. Notably, given a long context sentence in the QA task, LLM fails to utilize the context described in the middle~\cite{liu2023lost}. To handle the positional bias in the context-given QA task, re-ordering the input context~\cite{peysakhovich2023attention,jiang2023longllmlingua} or advanced training scheme~\cite{an2024make} is proposed. These work discuss the positional bias \wrt the contexts \textit{given as a prompt} during inference. In contrast, our interest is in the bias existing in knowledge of the training documents.

\textbf{Objectives in language modeling.}
Several works study the diverse denoising objectives and language modeling~\cite{wang2022language,tay2022unifying}. However, they do not investigate the positional bias caused by them. Our findings suggest that the nature of auto-regressive pre-training, \ie, a token is generated by seeing all previous tokens, causes this positional bias. Also, our work indicates that diverse regularization techniques can mitigate the bias for the causal language model. 

\vspace{-3mm}
\section{Dataset}
\vspace{-3mm}
We explain two datasets to evaluate LLM's ability to learn new knowledge and analyze the positional bias issue. We focus on whether fine-tuned LLM can answer questions about factual knowledge obtained from training documents. We leave most details for the appendix due to the limited space. First, we generate the synthetic biography dataset, following~\cite{zhu2023physics}, in an almost completely controlled manner. Second, we introduce a real dataset collected from the articles of Wikipedia, following~\cite{jiang2024instruction}, to study LLM's ability to adapt to the new domain with real documents. To assess the ability of LLMs to learn knowledge from new documents, we use a document corpus with minimal overlap with the original pre-training corpus. Since our goal is to investigate the positional bias in trained LLMs, our QA data is annotated with the corresponding source sentence from the documents. We also experiment on MedQuAD~\cite{medquad} by generating new QA pairs. See Sec.~\ref{sec:dataset_detail} for details.
\vspace{-2mm}
\subsection{Synthetic Bio}
\vspace{-3mm}
\textbf{Documents.} 
Following \cite{zhu2023physics}, we generate synthetic biography datasets consisting of sentences describing nine properties (birthday, birthplace, school, major, company, occupation, food, sports, and hobby) for 3000 individuals, where we ask ChatGPT for potential entries of each attribute and then randomly assign one attribute to each individual. Subsequently, we fit the properties in a sentence template, employing the same template for all persons. 
Consequently, the birthday is presented first, while the hobby is positioned last in each individual's description. If LLMs are free from positional bias, fine-tuned models can accurately answer all questions. See Fig. \ref{fig:synth_example} for an example of this dataset. 

\textbf{Questions.} 
We focus on evaluating the model's performance on the ability to retrieve nine properties. Note that we utilize the same question template for all individuals during inference. We randomly pick 500 persons for validation and testing respectively, and use the rest for training data. In total, 18000 questions are used for training and 4500 for validation and testing, respectively.

\subsection{Wiki2023+}
\vspace{-3mm}
\textbf{Documents.} 
We closely follow \cite{jiang2024instruction} to create the dataset, using Wikipedia 5911 articles classified under the “2023” category including topics from films, manga, sports, etc, to minimize the overlap with the pre-training corpus. 
We utilize only the summary section of the articles, which includes diverse factual information, to accelerate the training. The left of Fig.~\ref{fig:wiki2023} describes the document, “The Tenant (2023 film)”. Refer to the appendix for the details of this dataset. 

\textbf{Questions.} 
Following \cite{jiang2024instruction}, we employ LLM\footnote{We utilize GPT-3.5 turbo.} to generate the question-answer pairs from the article. While they generate questions by inputting an entire article into the LLM, we take a different approach by feeding each sentence individually to the LLM. This allows us to identify the source of each question. Consequently, our QA dataset contains annotations specifying the sentence responsible for generating each question, which eases the analysis of positional bias for this dataset. Since some generated QA pairs are inappropriate, \eg, some answers hallucinate information not described in the input article, or questions are irrelevant to the article, we filter QA samples to maintain the dataset's quality (See Sec.\ref{sec:qa_create} for details). After filtering, we assess the quality of randomly chosen QA pairs and estimate the overall quality, indicating that around 95\% of QA pairs are valid. 

\textbf{Data split.} 
We employ the domain of \textit{film} for our evaluation and randomly choose 1785, 100, and 500 documents, for training, validation, and testing respectively. Then, we train models on all 2385 film documents and QA data from 1785 training documents, and validation and testing are performed on each split\footnote{The numbers of QA pairs are 5493, 315, 1590 for training, validation, and testing respectively.}. This dataset includes 3526 articles from other genres for future investigation. 

\textbf{Skewed question distributions.}
The right of Fig.~\ref{fig:wiki2023} illustrates the histogram of the answer position in documents for the \textit{film} test split, revealing that the distribution of the answer position is skewed towards the beginning of the documents. Therefore, averaging the results on whole QA pairs does not reveal the robustness to the position of the answers. When benchmarking performance on this dataset in Sec.~\ref{sec:exp}, we introduce the evaluation metric to consider the skewness. 


\section{Experiments}\label{sec:exp}
\vspace{-3mm}
The goals of this section are twofold: (i) investigating the positional bias problem and (ii) benchmarking the performance of models on \fontchange{pcr}{Wiki2023+}. 
Sec.~\ref{sec:preliminary} describes the preliminary and Sec.~\ref{sec:recipes} explains the studied techniques. In Sec.~\ref{sec:answer_position}, we study the positional bias problem by modulating the position of the answer sentence in documents. In Sec.~\ref{sec:wiki2023}, we focus on evaluating the model's performance on \fontchange{pcr}{Wiki2023+} to provide a benchmark for continued training of LLMs. 

\subsection{Preliminary}\label{sec:preliminary}
\textbf{Pre-trained I-LLM.} 
Unless otherwise specified, we employ the open-sourced Llama-2 Chat model~\cite{touvron2023llama2}. Due to the limitation of computational resources, we mainly employ 7B for evaluation while providing analysis on 13B and 70B models.

\textbf{Optimization.}
We employ Adam\footnote{We also test on AdamW, but do not see a clear advantage over Adam.}, set the initial learning rate as 1e-5 with a linear decay scheduling, and train all models for 3000 steps, which correspond to 100 epochs in \fontchange{pcr}{Wiki2023+}, unless otherwise specified. Following~\cite{zhu2023physics}, we employ mixed sampling from QA and document data. 
Each mini-batch, 256 samples in total, randomly samples QA and document data. 

\textbf{Objective.}
To exploit the ability of the instruction-tuned Llama-2 model, we employ special tokens used in the Llama-2 Chat model for question-answer data. 
Specifically, each question sentence is wrapped with a “<INST></INST>” tag as shown in Fig.~\ref{fig:wiki2023}  where tokens used as prompts are highlighted. 
For the training loss, we compute the average negative log-likelihood loss only on tokens in the answer, $\bm{a}$, given the question, $\bm{q}$, $-\frac{1}{|\bm{a}|}\sum_{k}\log P(\bm{a}_{k}|\bm{q}, \bm{a}_{<k})$. 
Similarly, for the document data, we prompt the document, $\bm{d}$, using its title, $\bm{t}$, and compute the standard next-token prediction loss by averaging over all tokens in the document, $-\frac{1}{|\bm{d}|}\sum_{k}\log P(\bm{d}_{k}|\bm{t}, \bm{d}_{<k})$. 

\begin{figure}
    \centering
\includegraphics[width=\textwidth]{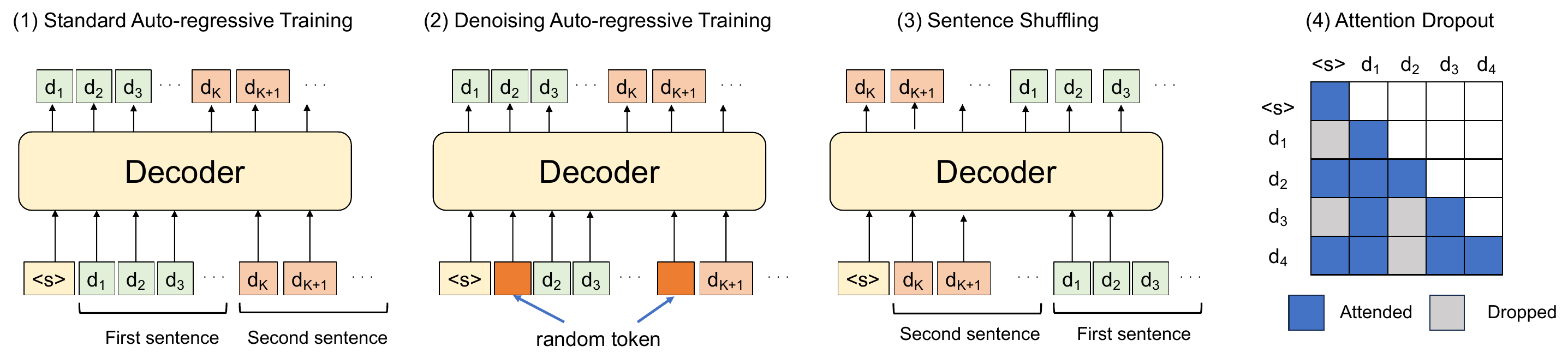}
    \caption{Visualization of studied four training methods. From left to right, (1) \textbf{AR}: standard auto-regressive training, (2) \textbf{D-AR}: denoising auto-regressive training randomly replaces input tokens with random ones while keeping the prediction target, (3) \textbf{Shuffle}: sentence shuffling shuffles input sentences, (4) \textbf{Attn Drop}: attention dropout randomly drops the attention in the self-attention module. }
    \label{fig:train_objectives}
\end{figure}

\textbf{Evaluation metric.}
Since most questions are simple and answers are short, we use exact-match
(EM) as our main metric~\cite{kwiatkowski2019natural}, which measures whether the model’s output matches the ground-truth answer exactly after normalization. To consider the longer answer and question, we also employ F-1, which considers both recall and precision of the answer. Although \fontchange{pcr}{bioS} 's answer is evenly distributed among all positions, \fontchange{pcr}{Wiki2023+} is highly skewed as shown in Fig.~\ref{fig:wiki2023}. In Sec~\ref{sec:wiki2023}, to benchmark the performance on \fontchange{pcr}{Wiki2023+}, we propose to compute the metric considering the location of the answers within documents. Specifically, we group the test QA pairs according to the position of the sentence generating the QA. Considering the number of QA pairs in each group, we group all positions of more than five into a sixth group, having six groups in total. Evaluation in each position reveals the robustness to the answer position. 

\subsection{Analyzed Training Recipes}\label{sec:recipes}
We hypothesize that excessive reliance on the previous tokens causes the “perplexity curse” and two factors are involved; (i) each factual knowledge is described by a single format, and (ii) vanilla auto-regressive training predicts the next token relying on all previous tokens. Suppose a sequence of three tokens or chunks of tokens, $A \rightarrow B \rightarrow C$, meaning subject $A$ does $B$, resulting in $C$. The auto-regressive model learns to prompt $C$ given \textit{fixed} $A$ and $B$, ensuring that $C$ is made extractable given the specific tokens $A$ and $B$. But, at test time, different expressions for $A$ and $B$ are often employed to extract information from $C$. Then, diversifying the expressions of tokens should generalize the connection among $A$, $B$, and $C$, easing the information extraction from diverse prompts. To address the issue, simpler techniques are more desirable considering LLM's training cost, and we study two data feeding methods and attention dropout as illustrated in Fig.~\ref{fig:train_objectives}. Note that we do not intend to claim the novelty of these techniques, instead, we highlight that these techniques have not been studied from better knowledge memorization and extraction.

\textbf{Denoising auto-regressive training (D-AR).} 
A natural option is to diversify the textual representations of a document, but such an approach requires a model good at translating documents into different expressions. 
As a simple and easy-to-plug-in method, we study replacing some input tokens with random ones, which perturbs $A$ or $B$ to prompt $C$ during training. Specifically, the training data generator chooses $R$\% of the token positions randomly and replaces the input token with a random one while the prediction loss is computed with original labels. The modified objective is written as $-\frac{1}{|\bm{d}|}\sum_{k}\log P(\bm{d}_{k}|\bm{\tilde{t}}, \bm{\tilde{d}}_{<k})$, where $\bm{\tilde{t}}$ and $\bm{\tilde{d}}_{<k}$ indicate the corrupted tokens. We think the reconstruction of the corrupted tokens does not contribute to the performance gain. More importantly, adding the noise into the input sequence enhances the model to predict the next token with diverse conditions, encouraging robust information extraction during testing (See Sec.~\ref{sec:appendix_exp}).
In the framework of masked language modeling, BERT~\cite{devlin2018bert} replaces some proportions of tones with random ones. We investigate the technique for \textit{auto-regressive training} to improve knowledge extraction ability. 

\textbf{Shuffling sentences (Shuffle).} 
Inspired by an approach~\cite{ko2020look} which tackles positional bias in question prompt, we study another simple remedy that shuffles the order of sentences in a document. If the model sees a different order of documents every time during training, the model will not rely on previous sentences to predict the next token. But, this remedy has two potential risks: (i) shuffling sentences can destroy the context from the previous sentence and (ii) this does not mitigate the spurious correlation within a single sentence. (i) can be critical if consecutive sentences describe a single fact, \eg, an explanation about some procedures, though sentences in our datasets are often independent. We leave the investigation on more advanced datasets for future work. 

\textbf{Attention dropout (Attn Drop).} We further study the regularization by dropout~\cite{hinton2012improving}. To mitigate the dependency on the previous tokens, we apply attention dropout, \ie, randomly dropping the attention mask, which should force the model to reduce the excessive reliance on previous tokens\footnote{This technique is widely used in pre-training language models though LLama-2 does not use it.}.


\subsection{Modulating the Position of Answer Sentence}\label{sec:answer_position}
We first study the effect of position in documents by modulating the position of the answer sentence, inspired by \cite{liu2023lost}. Suppose we have a document consisting of $n$ sentences, $\bm{D} = [\bm{s_1}, \bm{s_2}, ...\bm{s_n}]$, where $\bm{s_i}$ indicates a sentence. We evaluate the accuracy of answering a question about $\bm{s_1}$ by training a model on a set of documents, $\bm{D}^{k}$, which inserts $\bm{s_1}$ into $k$-th position, \eg,  $\bm{D}^{3} =  [\bm{s_2}, \bm{s_3}, \bm{s_1}...\bm{s_n}]$\footnote{Note that we perform this modulation on all articles in the dataset and tune models on them.}. This assesses the model's ability to memorize and retrieve information from $\bm{s_1}$ in different positions. 

\begin{figure}
\centering
\begin{subfigure}{0.47\textwidth}
  \centering
  \includegraphics[width=0.75\linewidth]{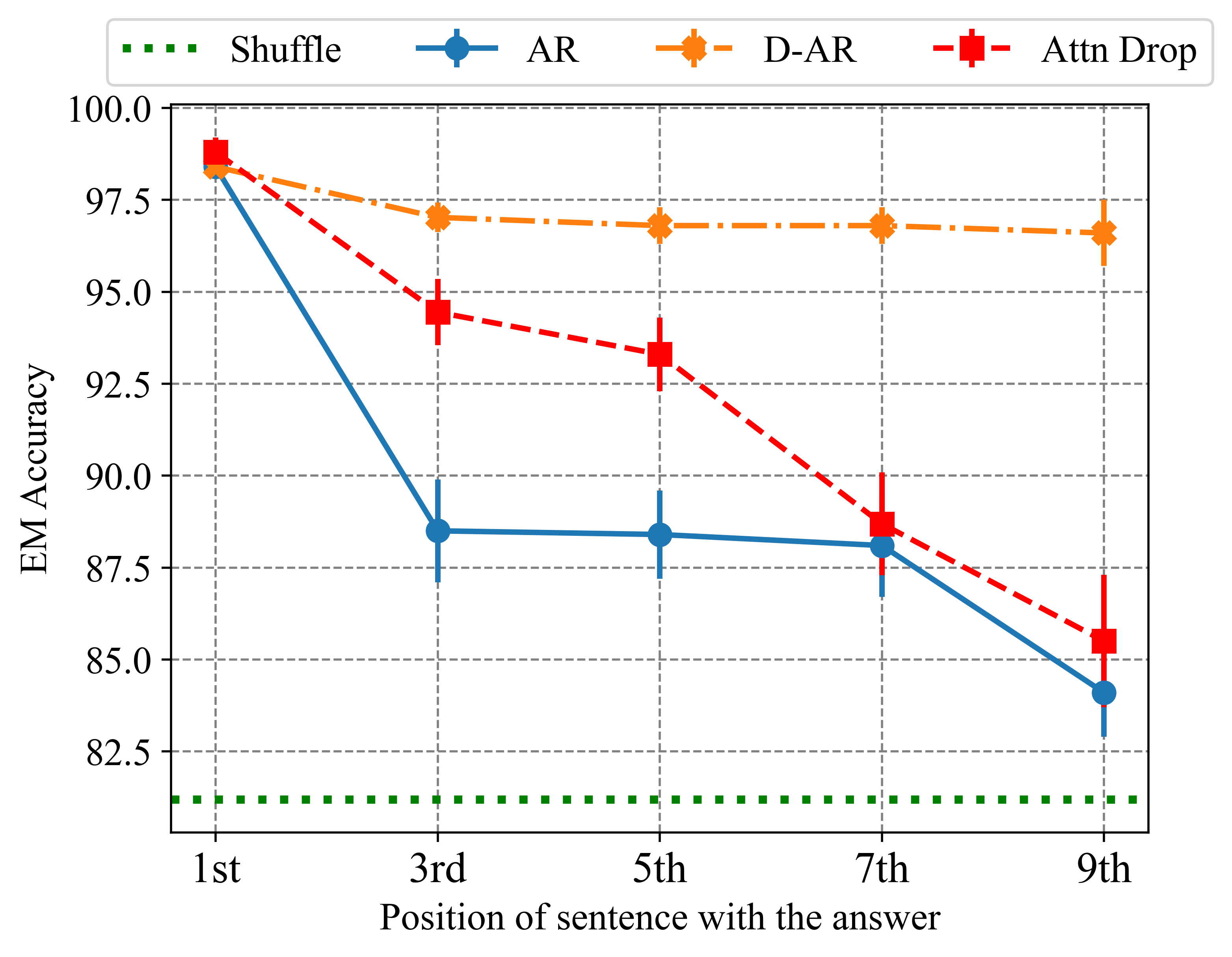} 
   \caption{EM accuracy on \fontchange{pcr}{bioS}.}
  \label{fig:em_bios}
\end{subfigure}%
\begin{subfigure}{0.45\textwidth}
  \centering
  \includegraphics[width=0.75\linewidth]{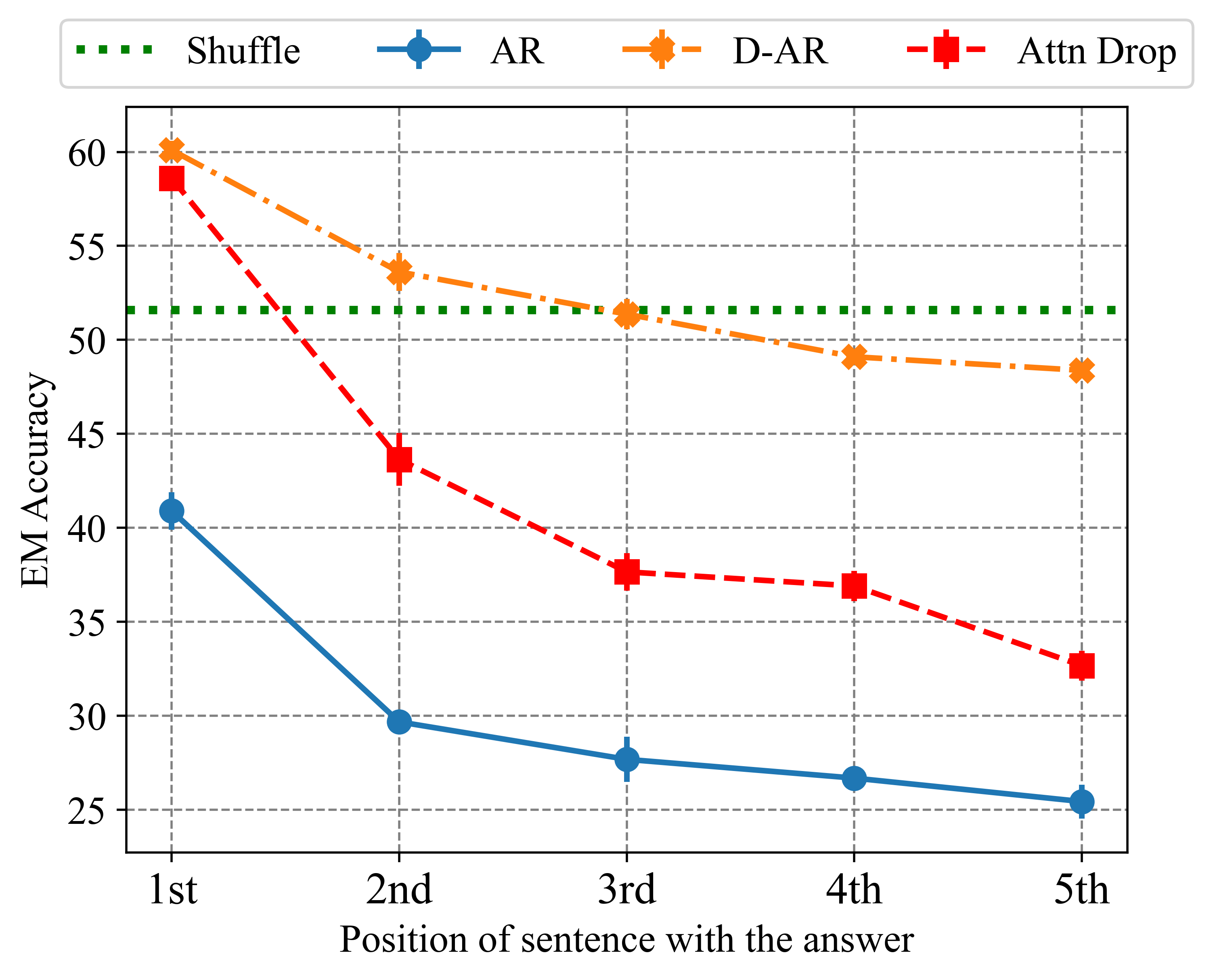}
   \caption{EM accuracy on \fontchange{pcr}{Wiki2023+}.}
  \label{fig:em_wiki}
\end{subfigure}
\caption{We vary the position of the sentence in training documents and fine-tune a model on the modulated documents. 
The X-axis represents $k$, the position of the sentence corresponding to the answer in the training documents ($\bm{D}^{k}$). Each point corresponds to a model trained on $\bm{D}^{k}$. 
\textbf{Left}: EM accuracy \wrt person's birthday for \fontchange{pcr}{bioS}. We vary the position of the sentence describing a person's birthday.  \textbf{Right}: EM accuracy on \fontchange{pcr}{Wiki2023+}. The position of the sentence corresponding to the answer varies from the 1st to the 5th.}
\label{fig:summary_result}
\end{figure}

\textbf{Vanilla AR model significantly suffers from positional bias.} 
In Fig. ~\ref{fig:summary_result}, we illustrate the results \wrt the position of the answer on \fontchange{qcr}{bios} and \fontchange{qcr}{Wiki2023+}. In both datasets, vanilla AR training significantly decreases the performance on the answer in the middle or end. 
Attn Drop improves the performance over AR but still suffers from the performance drop by the answer position. D-AR shows high performance in most positions, and the decrease by the position is limited compared to the AR and Attn Drop. Shuffle shows the lowest accuracy in \fontchange{qcr}{bios} probably because the model suffers from underfitting. Applying the regularization improves performance even in the first position for \fontchange{qcr}{Wiki2023+}, which is probably because of the effect from the answer position \textit{within a sentence}. If the answer is at the end of a long sentence, retrieving the information during inference can be hard because of the positional effect. In summary, this result indicates that the fine-tuned LLM struggles to retrieve information from the middle or end of a training document while regularization techniques such as D-AR and Attn Drop mitigate the issue in diverse positions. 

\begin{figure}
\centering
\begin{subfigure}{0.21\textwidth}
  \centering
  \includegraphics[width=0.8\linewidth]{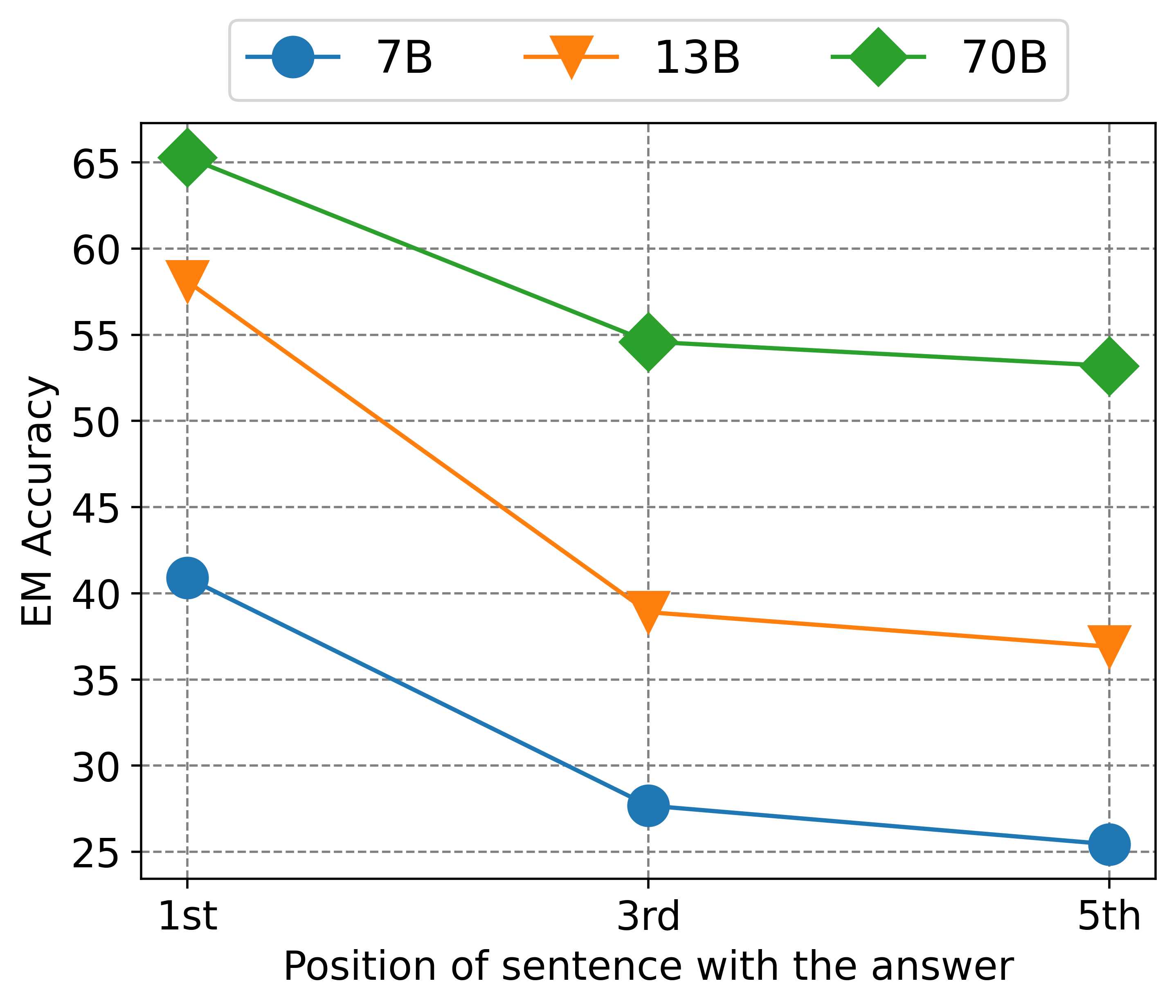} 
   \caption{7B vs. 13B vs. 70B.}
  \label{fig:compare_models}
\end{subfigure}%
\begin{subfigure}{0.19\textwidth}
  \centering
  \includegraphics[width=0.9\linewidth]{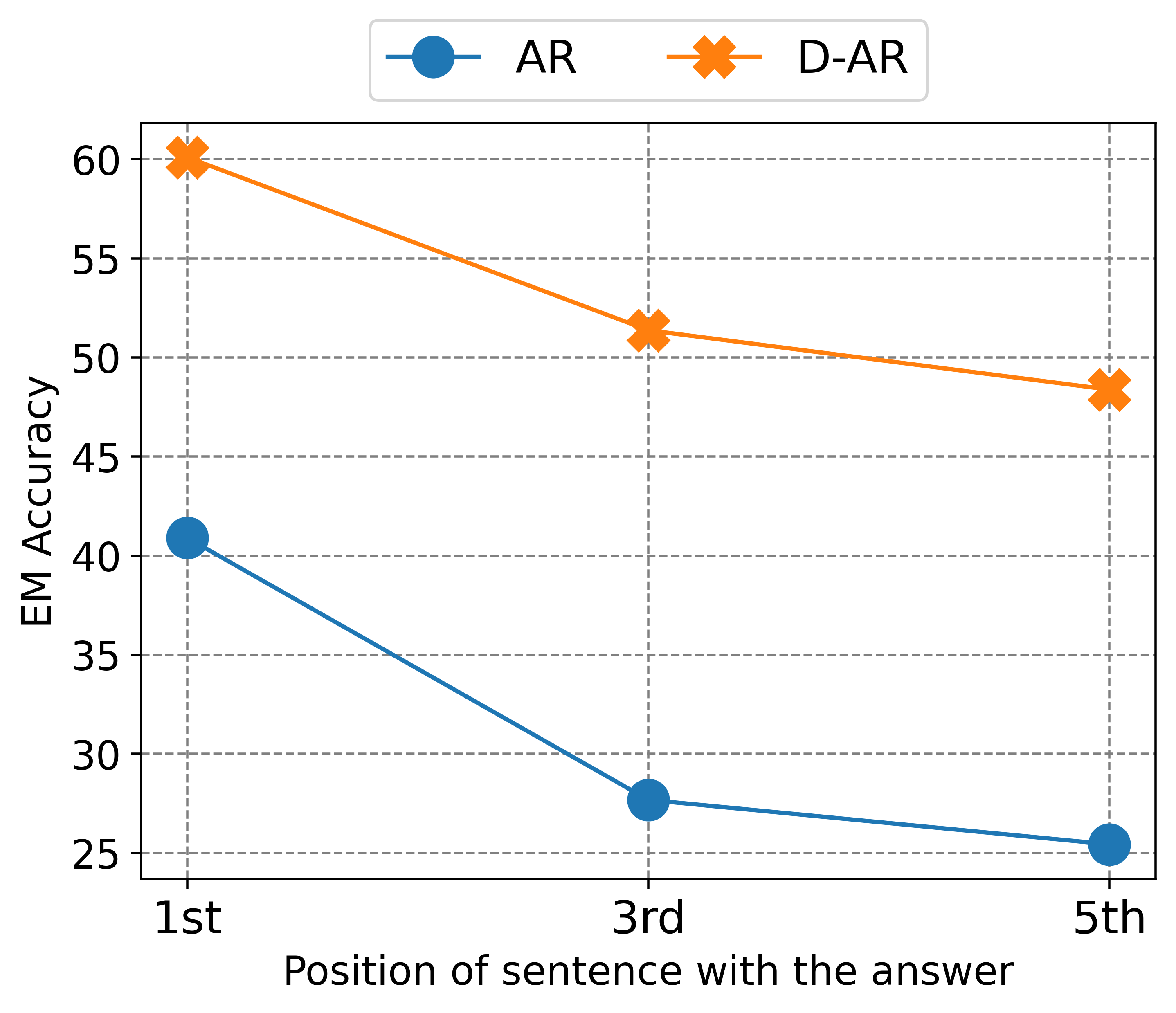}
   \caption{7B}
  \label{fig:7B}
\end{subfigure}
\begin{subfigure}{0.19\textwidth}
  \centering
  \includegraphics[width=0.9\linewidth]{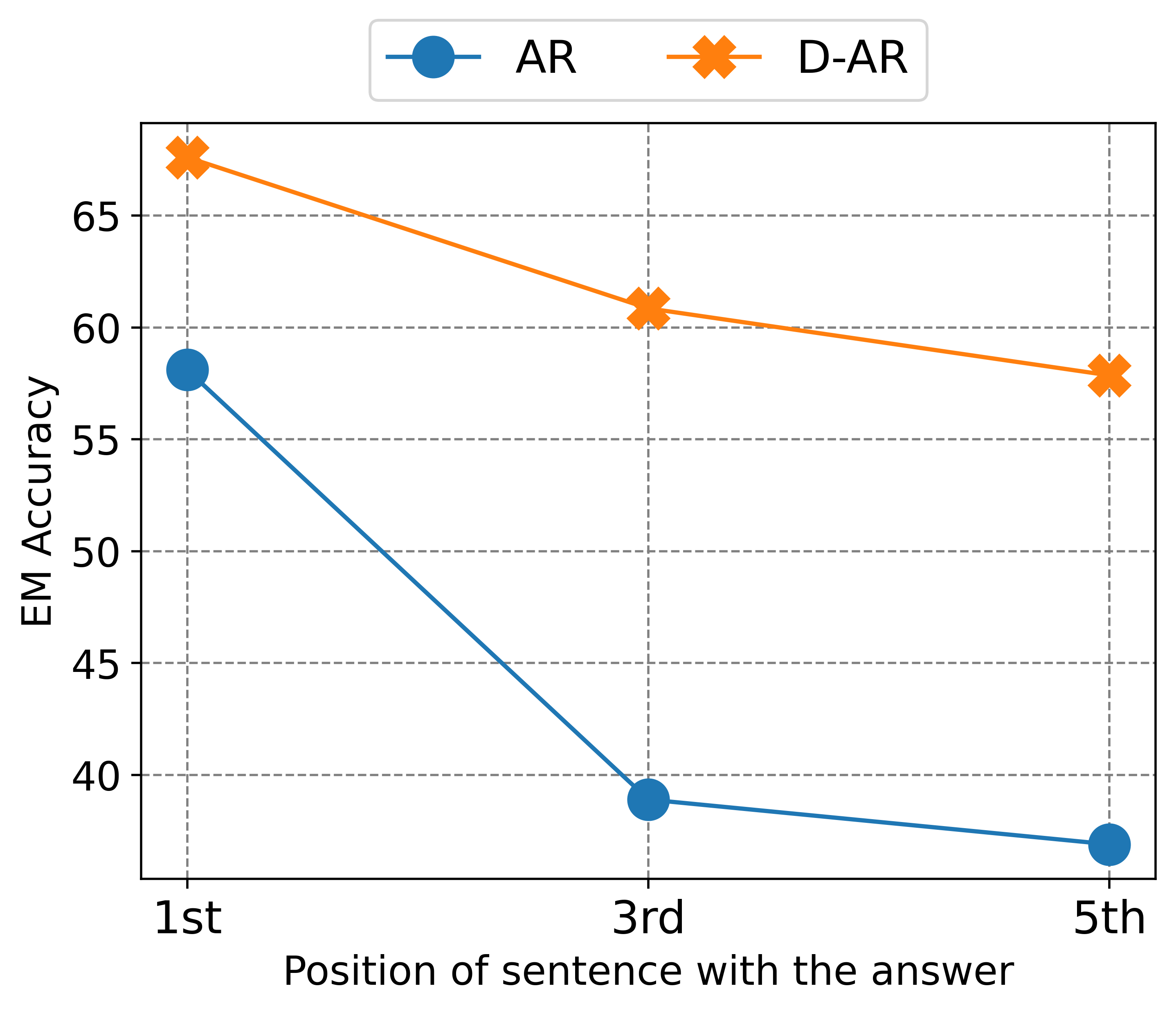}
   \caption{13B}
  \label{fig:13B}
\end{subfigure}
\begin{subfigure}{0.19\textwidth}
  \centering
  \includegraphics[width=0.92\linewidth]{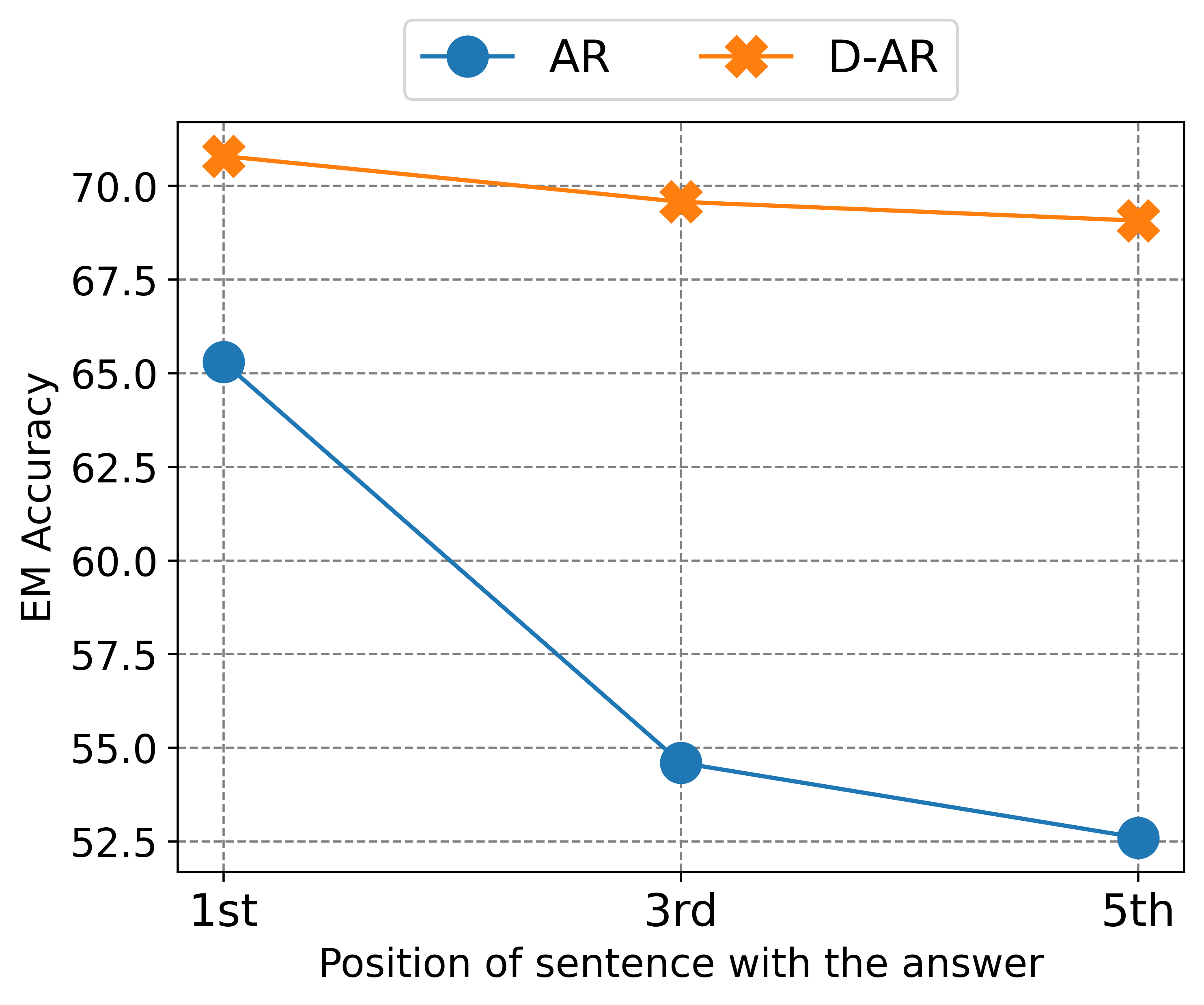}
   \caption{70B}
  \label{fig:70B}
\end{subfigure}
\begin{subfigure}{0.19\textwidth}
  \centering
  \includegraphics[width=0.92\linewidth]{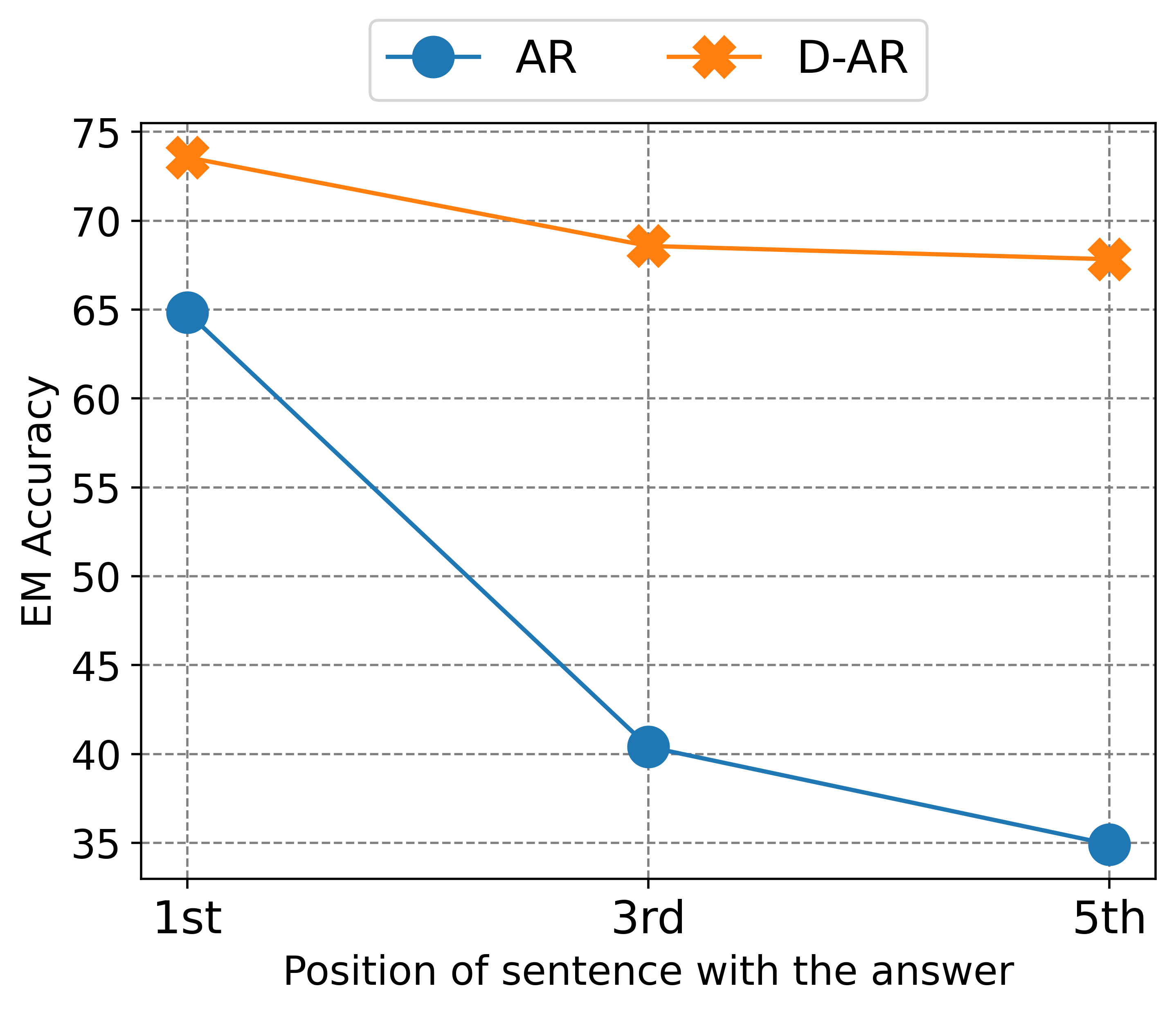}
   \caption{Zephyr-7B}
  \label{fig:zephyr}
\end{subfigure}
\caption{(a): Results of AR model with different sizes in modulating answer positions. We compare Llama-7B, 13B, and 70B for AR models. Models with increased parameter size improve performance, yet still suffer from the positional bias issue. (b)(c)(d): AR vs. D-AR in different model sizes. D-AR significantly improves performance over AR for all sizes. 70B model with D-AR greatly mitigates the effect from the answer position.}
\label{fig:model_size_analysis}
\end{figure}

\textbf{Strong models with regularization perform well.} Fig.~\ref{fig:model_size_analysis} studies larger models, \ie, LLama-2 13B and 70B. In Fig.~\ref{fig:compare_models}, larger AR models demonstrate better performance in all positions, yet they still significantly degrade the performance in the middle and end. In Fig.~\ref{fig:7B}, \ref{fig:13B} and \ref{fig:70B}, D-AR significantly boosts performance in all cases, and the D-AR 13B model outperforms the AR 70B model in all positions. For the D-AR 70B model, the performance decrease by the answer position is less than 2\%. 
The performance degradation of Zephyr-7B~\cite{tunstall2023zephyr} in Fig.~\ref{fig:zephyr} is approximately 5\%, and the model demonstrates high performance in all positions\footnote{Since Zephyr-7B was released on Sep. 2023, the training data can be leaked. However, the accuracy of Zephyr-7B, which is measured by Chat-GPT, is 7.7\% on this dataset, and we conclude that the model does not memorize the knowledge well in the pre-training stage despite the potential data leak.}. These indicate that a key to mitigating the “perplexity curse” is fine-tuning a strong model with a proper regularization method. 

\begin{figure}[t]
\centering
\begin{subfigure}{0.3\textwidth}
  \centering
   \includegraphics[width=0.98\linewidth]{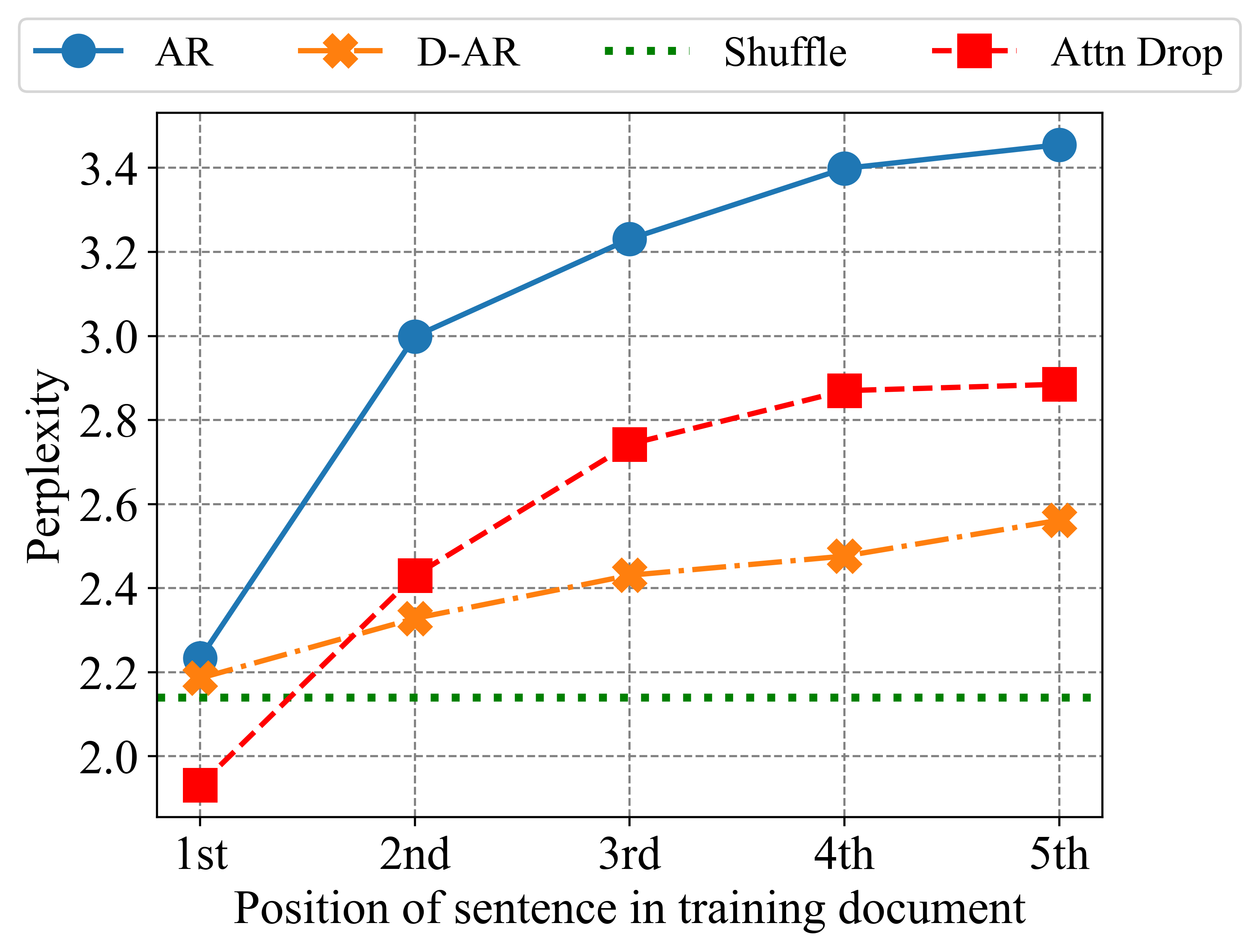} 
   \caption{Perplexity: LLama-7B.}
  \label{fig:perplexity_per_sentence}
\end{subfigure}%
\begin{subfigure}{0.3\textwidth}
  \centering
   \includegraphics[width=0.9\linewidth]{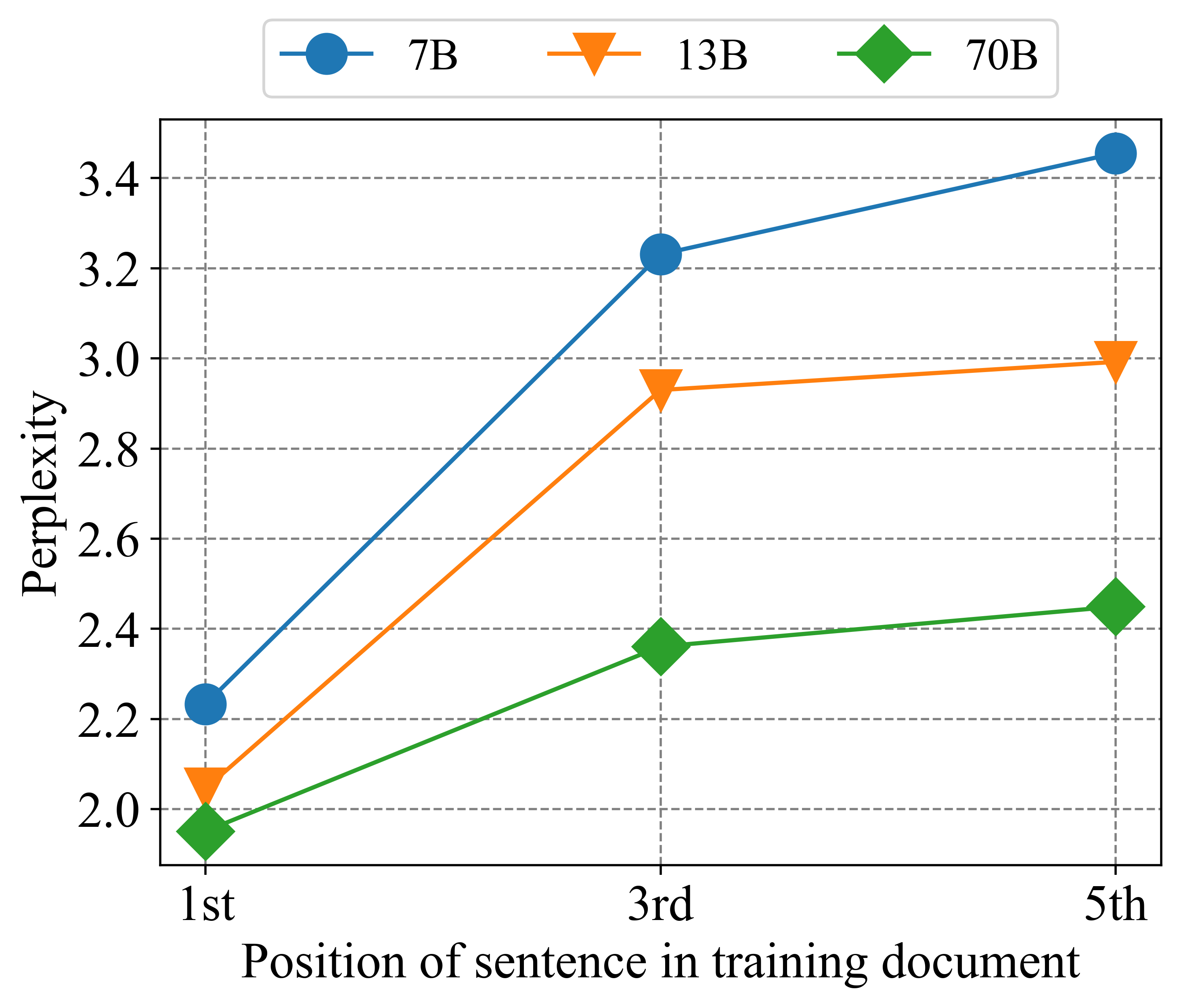} 
   \caption{Perplexity: 7B, 13B, 70B}
  \label{fig:perplexity_per_sentence_model}
\end{subfigure}%
\begin{subfigure}{0.3\textwidth}
  \centering
  \includegraphics[width=0.95\linewidth]{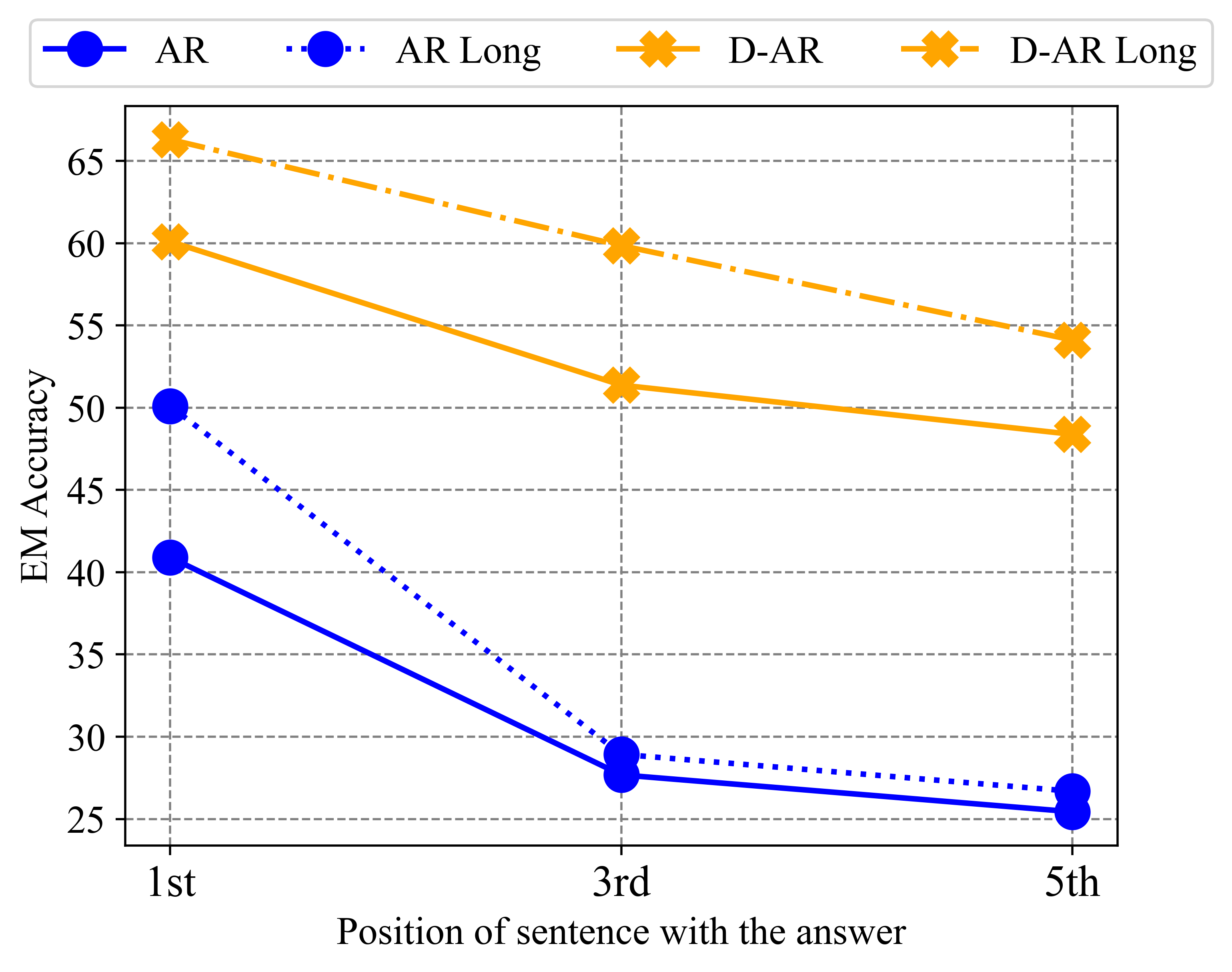}
   \caption{Longer training.}
  \label{fig:wiki_longer_training}
\end{subfigure}
\caption{\textbf{Left}: Analysis of perplexity conducted on models in Fig. ~\ref{fig:em_wiki}. We compute the perplexity for the first sentence in the original document ($\bm{s_1}$ in Sec.\ref{sec:answer_position}). The perplexity increases by putting the sentence latter for all models. AR model shows the highest perplexity. \textbf{Middle}: Perplexity comparison by model size. The smaller model relies more on the previous sentences to memorize a sentence. \textbf{Right}: Longer training benefits the performance improvements near the beginning of documents. Both analyses are conducted on \fontchange{pcr}{Wiki2023+}.}
\label{fig:perplexity_long_train}
\end{figure}

\textbf{Vanilla AR model memorizes documents in a hard-to-extract manner.} We use perplexity to study why the AR model struggles to retrieve information beyond the first sentence. First, the perplexities measured on $\bm{D}^{k}$ of \fontchange{pcr}{Wiki2023+} are almost 1.00 for all models, which indicates that the AR model memorizes the sentences almost perfectly given previous tokens. Next, we measure the perplexity of the first sentence in the original document, $\bm{s_1}$, for models trained with $\bm{D}^{k}$, \ie, models in Fig. ~\ref{fig:summary_result}. The models must remember each sentence without relying on previous sentences to minimize the perplexity. As shown in Fig.~\ref{fig:perplexity_per_sentence}, perplexity increases as putting the $\bm{s_1}$ latter for all models. This trend is the most notable for the AR model, implying that the AR model relies much on the previous sentences to remember each token. Therefore, appropriate information cannot be retrieved given the query sentence. 
The downward trend in perplexity is similar to the trend in QA in Fig.~\ref{fig:summary_result}. Figure ~\ref{fig:perplexity_per_sentence_model} analyzes the perplexity by model size, showing that larger models show smaller perplexity as reflected by QA performance. We do not aim to study whether perplexity can be a good measurement for models' behavior as done in \cite{xia2023training,tay2022scale}, but this analysis implies that the perplexity measured on training document ($\bm{D}^{k}$) does not reflect the knowledge extraction performance on diverse positions. 

\textbf{Longer training does not solve “perplexity curse”.} In Fig.~\ref{fig:wiki_longer_training}, we double the number of training iterations. 
For the AR model, increasing the training iterations improves the performance in the first position, but the improvement is limited in others, while the D-AR model improves in all positions. This indicates two facts: (i) increasing training iterations does not necessarily lead to the \textit{extractable} knowledge memorization, and (ii) regularization is important to benefit from longer training.

\begin{table*}[t]
\centering
\newcolumntype{Y}{>{\raggedright\arraybackslash}X}

\caption{EM and F1 score for each position of the answer in \fontchange{pcr}{Wiki2023+}. Compared to the auto-regressive model (AR), all techniques improve the knowledge extraction performance in all positions.}
\vspace{-2mm}
\scalebox{0.9}{
\begin{tabular}{c|cccccc|c}
\toprule
\multirow{2}{*}{Training}&\multicolumn{6}{c|}{$\longleftarrow$ start------------------------------------------------------------end$\longrightarrow$}&\multirow{2}{*}{Avg.}\\
&EM$_1$ / F1$_1$  &EM$_2$ / F1$_2$ &EM$_3$ / F1$_3$&EM$_4$ / F1$_4$&EM$_5$ / F1$_5$&EM$_6$ / F1$_6$&     \\\hline
AR&40.9 / 54.0 & 6.3 / 20.5&8.1 / 29.8&11.7 / 35.7&11.6 / 37.8 &10.7 / 36.4&14.9 / 35.7\\
Shuffle & 51.6 / 65.7& 14.7 / 43.2& 15.6 / 43.5& 20.6 / 46.8& 24.0 / 50.8&19.8 / 46.4&24.4 / 49.4\\
Attn Drop & 58.6 / 71.1 & 10.2 / 29.8 & 14.0 / 36.6 & 17.0	/ 38.6 & 13.2 / 42.8 & 13.3 / 39.7	&21.0 / 43.1 \\
D-AR&\textbf{60.1} / \textbf{73.7}& \textbf{26.9} / \textbf{53.1}& \textbf{23.4} / \textbf{52.9}& \textbf{26.0} / \textbf{51.7} & \textbf{24.8} / \textbf{52.2}& \textbf{21.3} / \textbf{48.2}&\textbf{30.4} / \textbf{55.3}\\

\bottomrule
\end{tabular}}
\label{tab:position_wiki}
\end{table*}

\subsection{Benchmark on unmodulated \fontchange{qcr}{Wiki2023+}}\label{sec:wiki2023}
We aim to benchmark the performance of models with different regularization techniques on \fontchange{qcr}{Wiki2023+}. Unlike Sec.~\ref{sec:answer_position} computing the accuracy only for the first sentence, $\bm{s_1}$, we train all models on the \textit{unmodulated} documents and compute metrics for each sentence position. 

\textbf{Overview of results on \fontchange{qcr}{Wiki2023+}.} 
Table~\ref{tab:position_wiki} details the results on \fontchange{qcr}{Wiki2023+}. D-AR significantly improves performance over the AR model. 
All regularization techniques improve the performance in all positions over the AR model. 
Only positional bias cannot explain the low performance in the second to sixth positions. We hypothesize that the content described in the latter sentences can be harder to answer; the first sentence often explains the simple fact, \eg, who made the film or when it was made, while the latter sentences can contain more complex facts. 


\begin{table*}[t]
\centering
\newcolumntype{Y}{>{\raggedright\arraybackslash}X}

\caption{Comparison by model size in \fontchange{pcr}{Wiki2023+}. Larger models still struggle to retrieve information from the end of documents. D-AR significantly improves performance in all positions.}
\scalebox{0.83}{
\begin{tabular}{cc|cccccc|c}
\toprule
\multirow{2}{*}{Model Size}&\multirow{2}{*}{Method}&\multicolumn{6}{c|}{$\longleftarrow$ start------------------------------------------------------------end$\longrightarrow$}&\multirow{2}{*}{Avg.}\\
&&EM$_1$ / F1$_1$  &EM$_2$ / F1$_2$ &EM$_3$ / F1$_3$&EM$_4$ / F1$_4$&EM$_5$ / F1$_5$&EM$_6$ / F1$_6$&     \\\hline
\multirow{2}{*}{7B}&AR&40.9 / 54.0 & 6.3 / 20.5&8.1 / 29.8&11.7 / 35.7&11.6 / 37.8 &10.7 / 36.4&14.9 / 35.7\\
&D-AR&\textbf{60.1} / \textbf{73.7}& \textbf{26.9} / \textbf{53.1}& \textbf{23.4} / \textbf{52.9}& \textbf{26.0} / \textbf{51.7} & \textbf{24.8} / \textbf{52.2}& \textbf{21.3} / \textbf{48.2}&\textbf{30.4} / \textbf{55.3}\\\hline
\multirow{2}{*}{13B}&AR&58.1 / 69.3 & 8.7 / 28.3 & 	18.8 / 40.2 & 20.2 / 42.3 & 11.6 / 39.2 & 14.7 / 39.3 & 22.0 / 43.1 \\
&D-AR&\textbf{67.6} / \textbf{84.1} & \textbf{34.4} / \textbf{64.4}& \textbf{32.8} / \textbf{64.0} & \textbf{30.5} / \textbf{58.6}& \textbf{30.2} / \textbf{59.0} & \textbf{22.8} / \textbf{52.2} & \textbf{36.4} / \textbf{63.7}\\\hline
\multirow{2}{*}{70B}&AR&65.3 / 78.9& 27.2 / 48.9& 24.4 / 46.2& 27.8 / 50.9& 22.5 / 50.9& 22.8 / 48.1& 31.7 / 54.0\\ 
&D-AR & \textbf{70.8} / \textbf{85.8} & \textbf{48.8} / \textbf{68.9} & \textbf{43.8} / \textbf{70.7} & \textbf{39.5} / \textbf{64.8} & \textbf{38.0} / \textbf{66.7} & \textbf{36.0} / \textbf{60.7} & \textbf{46.2} / \textbf{69.6}\\
\bottomrule
\end{tabular}}
\label{tab:modelsize_wiki}
\end{table*}
\textbf{Analysis of larger models.} 
Table~\ref{tab:modelsize_wiki} studies larger models. Larger models demonstrate better performance in all positions, yet they still significantly degrade the performance in the middle and end. D-AR significantly boosts performance in all cases. 


\begin{table*}[t]
\centering
\newcolumntype{Y}{>{\raggedright\arraybackslash}X}

\caption{Combining regularization techniques with denoising auto-regressive training (D-AR) in \fontchange{pcr}{Wiki2023+}. These methods complement each other, and combining them boosts performance.}
\scalebox{0.8}{
\begin{tabular}{cc|cccccc|c}
\toprule

\multirow{2}{*}{Shuffle}&\multirow{2}{*}{Attn Drop}&\multicolumn{6}{c|}{$\longleftarrow$ start------------------------------------------------------------end$\longrightarrow$}&\multirow{2}{*}{Avg.}\\
&&EM$_1$ / F1$_1$  &EM$_2$ / F1$_2$ &EM$_3$ / F1$_3$&EM$_4$ / F1$_4$&EM$_5$ / F1$_5$&EM$_6$ / F1$_6$&     \\\hline

&&60.1 / 73.7& 26.9 / 53.1& 23.4 / 52.9& 26.0 / 51.7 & 24.8 / 52.2& 21.3 / 48.2&30.4 / 55.3\\

\checkmark & & 59.9 / 76.8	& 19.5 / 60.7 & 26.3 / 61.0 & 32.7 / 62.6 &32.6 / 65.0	&	24.4 / 57.5& 32.5 / 63.9\\

& \checkmark & \textbf{68.3}	/ \textbf{81.9} & \textbf{32.1} / 59.3 & 29.2 / 60.5 & 27.4 / 53.7&  23.3 / 55.0 & 22.4 / 51.4 & 33.8 / 60.3\\

\checkmark &  \checkmark & 65.8 / 81.8& 24.9 / \textbf{66.5}& \textbf{29.5} / \textbf{64.9}& \textbf{37.7} / \textbf{67.9}& \textbf{33.3} / \textbf{67.1} & \textbf{24.5} / \textbf{59.6}& \textbf{36.0} / \textbf{68.0}\\

\bottomrule
\end{tabular}}
\label{tab:combine_reg}
\end{table*}
\textbf{Combining different regularization improves performance.} Given Table~\ref{tab:position_wiki} and Fig.~\ref{fig:summary_result}, D-AR is the most effective technique to mitigate positional bias. Then, we investigate the combination with D-AR and other regularization techniques in Table~\ref{tab:combine_reg}. Combining Shuffle and D-AR (second row) significantly improves F1, indicating the improvement in answering longer sequences. Also, Shuffle enhances the performance gain in the last three groups. Attn Drop tends to improve in the first three positions (third row). On average, combining all techniques (last row) produces a notable gain in both EM and F1. From these results, we conclude that the studied techniques can complement each other. 

\begin{table*}[t]
\centering
\newcolumntype{Y}{>{\raggedright\arraybackslash}X}

\caption{Effectiveness of adding paragraphs translated by ChatGPT in \fontchange{pcr}{Wiki2023+}. Adding translated documents improves performance, but the effectiveness on EM$_6$ is limited.}
\scalebox{0.8}{
\begin{tabular}{c|c|cccccc|c}
\toprule

\multirow{2}{*}{Method}&\multirow{2}{*}{Augment}&\multicolumn{6}{c|}{$\longleftarrow$ start------------------------------------------------------------end$\longrightarrow$}&\multirow{2}{*}{Avg.}\\
&&EM$_1$ / F1$_1$  &EM$_2$ / F1$_2$ &EM$_3$ / F1$_3$&EM$_4$ / F1$_4$&EM$_5$ / F1$_5$&EM$_6$ / F1$_6$&     \\\hline

\multirow{2}{*}{AR}&&40.9 / 54.0 & 6.3 / 20.5&8.1 / 29.8&11.7 / 35.7&11.6 / 37.8 &10.7 / 36.4&14.9 / 35.7\\

& \checkmark & \textbf{58.1} / \textbf{70.7} & \textbf{12.3} / \textbf{29.6} & \textbf{15.6} / \textbf{37.6} & \textbf{16.6} / \textbf{40.8} & \textbf{15.5} / \textbf{42.1} & \textbf{13.2} / \textbf{41.6} & \textbf{21.9} / \textbf{43.7} \\\hline

\multirow{2}{*}{D-AR}&&60.1 / 73.7& 26.9 / 53.1& 23.4 / 52.9& 26.0 / 51.7 & 24.8 / 52.2& 21.3 / 48.2&30.4 / 55.3\\ 
&\checkmark & \textbf{65.8}	/ \textbf{78.8} & \textbf{37.4} / \textbf{64.2} & \textbf{37.0} / \textbf{65.3} & \textbf{35.9} / \textbf{62.5}& \textbf{28.7} / \textbf{61.0} & \textbf{21.8} / \textbf{53.9} & \textbf{37.8} / \textbf{64.3}\\
\bottomrule
\end{tabular}}
\label{tab:augmentation}
\end{table*}

\textbf{Diversifying document data. }
\cite{zhu2023physics} demonstrate that diversifying the representations of each document can enhance the knowledge extraction ability. Diversifying the representations can encourage the model to elicit knowledge with different but the same meanings of queries, which should mitigate the positional bias. But, diversifying real-world documents' representations is not a trivial operation, and necessitates an accurate paragraph-to-paragraph translation model. To study the effectiveness, we utilize Chat-GPT to rephrase the documents of \fontchange{pcr}{Wiki2023+} in four ways using a prompt to diversify the order of sentences and train the model combined with the original documents. 
According to Table~\ref{tab:augmentation}, we have two observations: (i) the augmentation improves performance overall and is complementary to D-AR, and (ii) it significantly improves performance on the answers described near the beginning, but the effectiveness is limited on those near the end. For (ii), we qualitatively find that paragraphs generated by Chat-GPT do not change the sentence order much despite using prompts to diversity the order, which explains why improvements at the end are limited. 

\begin{wraptable}[7]{r}{0.4\linewidth}
    \centering
    \vspace{-3mm}
        \caption{Results on MedQuAD.}
        \vspace{-2mm}
    \scalebox{0.9}{
    \begin{tabular}{c|ccc}
        \toprule
      Method & EM & F1 & GPT-Eval\\\hline
      No tuning & 0.0 & 5.1 & 6.9\\
      AR & 8.9&35.9&26.6\\
      Shuffle & 5.6 & 29.2 & 21.7\\
      D-AR & 10.9&39.0&29.3\\
      Attn Drop & 10.6 & 38.1&29.1\\
        \bottomrule
    \end{tabular}}
    \label{tab:res_medquad}
\end{wraptable}
\textbf{Adaptation to medical domain.} 
We investigate the effectiveness of regularization in the adaptation to the medical domain, using MedQuAD~\cite{medquad} in Table~\ref{tab:res_medquad}. In addition to EM and F1, we compute the accuracy by asking ChatGPT if the predicted answer is correct given the question and ground-truth answer, shown as GPT-Eval~\cite{liu-etal-2023-g} (See Sec.~\ref{sec:details_exp} for details). Both D-AR and Attn Drop improve performance over AR. See Sec.~\ref{sec:dataset_detail} for the details. 

\textbf{Comparison with Open-book QA.} We report the performance where the corresponding document of \fontchange{pcr}{Wiki2023+} is provided with Llama-2 7B as a context. This setting assumes the RAG with a perfect retrieval model. 
Since its answer format differs a lot from \fontchange{pcr}{Wiki2023+}, applying EM or F1 evaluation metric is unfair. Then, we utilize ChatGPT for automatic evaluation, finding that the accuracy of the open-book setting is 90.6\% while that of the best model in Table~\ref{tab:modelsize_wiki} is 50.8 \%. This indicates that the RAG outperforms the continued training if it has an accurate retrieval model. 

\textbf{Discussion: RAG or fine-tuning.} 
There are three main disadvantages in RAG compared to fine-tuning: (i) the RAG shows high latency given a long context, (ii) needs a good retrieval model, and (iii) suffers from the lost-in-the-middle though many works attempt to address it~\cite{an2024make}. By contrast, the fine-tuning framework has three main disadvantages: (i) it must fine-tune LLMs to adapt to the new data, (ii) suffers from the positional bias as shown in our paper, and (iii) the amount of information storable in the model's parameters is limited. The effectiveness of each framework should depend on the application and the unification of two approaches can be desirable, \eg, fine-tuning a general knowledge learner on a new domain and complementing up-to-date information using RAG.

\vspace{-4mm}
\section{Conclusion}\label{sec:conclusion}
\vspace{-3mm}
In this work, we investigate the issue of “perplexity curse” in the continued training of LLM. Then, we find a very intriguing fact that LLMs struggle to extract information beyond the first sentence. Our study indicates that auto-regressive training forces the model to memorize contents relying on many irrelevant tokens, and simple and easy techniques such as denoising auto-regressive training and attention dropout mitigate the issue. We will release the code and dataset to reproduce our results, which should encourage more researchers to investigate the issue. 

\textbf{Limitations.} Our work has two limitations. First, fine-tuned LLM underperforms the Open-book QA baseline as discussed in Sec.~\ref{sec:exp}. However, our work provides an issue of auto-regressive loss and can be a key to closing the gap between the memorize-and-extract method and the Open-book QA approach. Second, we focus on QA for simple factual knowledge. We need to develop advanced datasets to study more complex knowledge acquisition scenarios.

\textbf{Acknowledgement.} We thank Shusaku Sone, Tatsunori Taniai, Jiaxin Ma, Donghyun Kim, and Chun-Liang Li for their valuable feedback on our research. This work is supported by JST Moonshot R\&D Program Grant Number JPMJMS2236. We used the computational resources of AI Bridging Cloud Infrastructure (ABCI) provided by National Institute of Advanced Industrial Science and Technology (AIST).

\bibliographystyle{unsrt}
\bibliography{arxiv_neurips}

\begin{thebibliography}{10}

\bibitem{touvron2023llama}
Hugo Touvron, Thibaut Lavril, Gautier Izacard, Xavier Martinet, Marie-Anne
  Lachaux, Timoth{\'e}e Lacroix, Baptiste Rozi{\`e}re, Naman Goyal, Eric
  Hambro, Faisal Azhar, et~al.
\newblock Llama: Open and efficient foundation language models.
\newblock {\em arXiv preprint arXiv:2302.13971}, 2023.

\bibitem{workshop2022bloom}
BigScience Workshop, Teven~Le Scao, Angela Fan, Christopher Akiki, Ellie
  Pavlick, Suzana Ili{\'c}, Daniel Hesslow, Roman Castagn{\'e}, Alexandra~Sasha
  Luccioni, Fran{\c{c}}ois Yvon, et~al.
\newblock Bloom: A 176b-parameter open-access multilingual language model.
\newblock {\em arXiv preprint arXiv:2211.05100}, 2022.

\bibitem{touvron2023llama2}
Hugo Touvron, Louis Martin, Kevin Stone, Peter Albert, Amjad Almahairi, Yasmine
  Babaei, Nikolay Bashlykov, Soumya Batra, Prajjwal Bhargava, Shruti Bhosale,
  et~al.
\newblock Llama 2: Open foundation and fine-tuned chat models.
\newblock {\em arXiv preprint arXiv:2307.09288}, 2023.

\bibitem{penedo2023refinedweb}
Guilherme Penedo, Quentin Malartic, Daniel Hesslow, Ruxandra Cojocaru,
  Alessandro Cappelli, Hamza Alobeidli, Baptiste Pannier, Ebtesam Almazrouei,
  and Julien Launay.
\newblock The refinedweb dataset for falcon llm: outperforming curated corpora
  with web data, and web data only.
\newblock {\em arXiv preprint arXiv:2306.01116}, 2023.

\bibitem{brown2020language}
Tom Brown, Benjamin Mann, Nick Ryder, Melanie Subbiah, Jared~D Kaplan, Prafulla
  Dhariwal, Arvind Neelakantan, Pranav Shyam, Girish Sastry, Amanda Askell,
  et~al.
\newblock Language models are few-shot learners.
\newblock In {\em NeurIPS}, 2020.

\bibitem{raffel2020exploring}
Colin Raffel, Noam Shazeer, Adam Roberts, Katherine Lee, Sharan Narang, Michael
  Matena, Yanqi Zhou, Wei Li, and Peter~J Liu.
\newblock Exploring the limits of transfer learning with a unified text-to-text
  transformer.
\newblock {\em The Journal of Machine Learning Research}, 2020.

\bibitem{gao2020pile}
Leo Gao, Stella Biderman, Sid Black, Laurence Golding, Travis Hoppe, Charles
  Foster, Jason Phang, Horace He, Anish Thite, Noa Nabeshima, et~al.
\newblock The pile: An 800gb dataset of diverse text for language modeling.
\newblock {\em arXiv preprint arXiv:2101.00027}, 2020.

\bibitem{wei2021finetuned}
Jason Wei, Maarten Bosma, Vincent~Y Zhao, Kelvin Guu, Adams~Wei Yu, Brian
  Lester, Nan Du, Andrew~M Dai, and Quoc~V Le.
\newblock Finetuned language models are zero-shot learners.
\newblock In {\em ICLR}, 2022.

\bibitem{han2023medalpaca}
Tianyu Han, Lisa~C Adams, Jens-Michalis Papaioannou, Paul Grundmann, Tom
  Oberhauser, Alexander L{\"o}ser, Daniel Truhn, and Keno~K Bressem.
\newblock Medalpaca--an open-source collection of medical conversational ai
  models and training data.
\newblock {\em arXiv preprint arXiv:2304.08247}, 2023.

\bibitem{wu2023pmc}
Chaoyi Wu, Xiaoman Zhang, Ya~Zhang, Yanfeng Wang, and Weidi Xie.
\newblock Pmc-llama: Further finetuning llama on medical papers.
\newblock {\em arXiv preprint arXiv:2304.14454}, 2023.

\bibitem{li2023llava}
Chunyuan Li, Cliff Wong, Sheng Zhang, Naoto Usuyama, Haotian Liu, Jianwei Yang,
  Tristan Naumann, Hoifung Poon, and Jianfeng Gao.
\newblock Llava-med: Training a large language-and-vision assistant for
  biomedicine in one day.
\newblock {\em arXiv preprint arXiv:2306.00890}, 2023.

\bibitem{jang2022towards}
Joel Jang, Seonghyeon Ye, Sohee Yang, Joongbo Shin, Janghoon Han, Gyeonghun
  KIM, Stanley~Jungkyu Choi, and Minjoon Seo.
\newblock Towards continual knowledge learning of language models.
\newblock In {\em ICLR}, 2022.

\bibitem{hu2023meta}
Nathan Hu, Eric Mitchell, Christopher~D Manning, and Chelsea Finn.
\newblock Meta-learning online adaptation of language models.
\newblock {\em arXiv preprint arXiv:2305.15076}, 2023.

\bibitem{allen2023physics}
Zeyuan Allen-Zhu and Yuanzhi Li.
\newblock Physics of language models: Part 1, context-free grammar.
\newblock {\em arXiv preprint arXiv:2305.13673}, 2023.

\bibitem{jiang2024instruction}
Zhengbao Jiang, Zhiqing Sun, Weijia Shi, Pedro Rodriguez, Chunting Zhou, Graham
  Neubig, Xi~Victoria Lin, Wen-tau Yih, and Srinivasan Iyer.
\newblock Instruction-tuned language models are better knowledge learners.
\newblock {\em arXiv preprint arXiv:2402.12847}, 2024.

\bibitem{zhu2023physics}
Zeyuan~Allen Zhu and Yuanzhi Li.
\newblock Physics of language models: Part 3.1, knowledge storage and
  extraction.
\newblock {\em arXiv preprint arXiv:2309.14316}, 2023.

\bibitem{liu2023lost}
Nelson~F Liu, Kevin Lin, John Hewitt, Ashwin Paranjape, Michele Bevilacqua,
  Fabio Petroni, and Percy Liang.
\newblock Lost in the middle: How language models use long contexts.
\newblock {\em TACL}, 2023.

\bibitem{mitchell2021fast}
Eric Mitchell, Charles Lin, Antoine Bosselut, Chelsea Finn, and Christopher~D
  Manning.
\newblock Fast model editing at scale.
\newblock In {\em ICLR}, 2022.

\bibitem{mitchell2022memory}
Eric Mitchell, Charles Lin, Antoine Bosselut, Christopher~D Manning, and
  Chelsea Finn.
\newblock Memory-based model editing at scale.
\newblock In {\em ICML}, 2022.

\bibitem{meng2022locating}
Kevin Meng, David Bau, Alex Andonian, and Yonatan Belinkov.
\newblock Locating and editing factual associations in gpt.
\newblock In {\em NeurIPS}, 2022.

\bibitem{meng2022memit}
Kevin Meng, Arnab Sen~Sharma, Alex Andonian, Yonatan Belinkov, and David Bau.
\newblock Mass editing memory in a transformer.
\newblock In {\em ICLR}, 2023.

\bibitem{feigenbaum2024editing}
Itai Feigenbaum, Devansh Arpit, Huan Wang, Shelby Heinecke, Juan~Carlos
  Niebles, Weiran Yao, Caiming Xiong, and Silvio Savarese.
\newblock Editing arbitrary propositions in llms without subject labels.
\newblock {\em arXiv preprint arXiv:2401.07526}, 2024.

\bibitem{wang2023knowledge}
Song Wang, Yaochen Zhu, Haochen Liu, Zaiyi Zheng, Chen Chen, et~al.
\newblock Knowledge editing for large language models: A survey.
\newblock {\em arXiv preprint arXiv:2310.16218}, 2023.

\bibitem{lewis2020retrieval}
Patrick Lewis, Ethan Perez, Aleksandra Piktus, Fabio Petroni, Vladimir
  Karpukhin, Naman Goyal, Heinrich K{\"u}ttler, Mike Lewis, Wen-tau Yih, Tim
  Rockt{\"a}schel, et~al.
\newblock Retrieval-augmented generation for knowledge-intensive nlp tasks.
\newblock In {\em NeurIPS}, 2020.

\bibitem{guu2020retrieval}
Kelvin Guu, Kenton Lee, Zora Tung, Panupong Pasupat, and Mingwei Chang.
\newblock Retrieval augmented language model pre-training.
\newblock In {\em ICML}. PMLR, 2020.

\bibitem{hofstatter2022multi}
Sebastian Hofst{\"a}tter, Jiecao Chen, Karthik Raman, and Hamed Zamani.
\newblock Multi-task retrieval-augmented text generation with relevance
  sampling.
\newblock {\em arXiv preprint arXiv:2207.03030}, 2022.

\bibitem{ko2020look}
Miyoung Ko, Jinhyuk Lee, Hyunjae Kim, Gangwoo Kim, and Jaewoo Kang.
\newblock Look at the first sentence: Position bias in question answering.
\newblock In {\em EMNLP}, 2020.

\bibitem{ma2021exploiting}
Fang Ma, Chen Zhang, and Dawei Song.
\newblock Exploiting position bias for robust aspect sentiment classification.
\newblock In Chengqing Zong, Fei Xia, Wenjie Li, and Roberto Navigli, editors,
  {\em ACL-IJCNLP}, 2021.

\bibitem{hofstatter2021mitigating}
Sebastian Hofst{\"a}tter, Aldo Lipani, Sophia Althammer, Markus Zlabinger, and
  Allan Hanbury.
\newblock Mitigating the position bias of transformer models in passage
  re-ranking.
\newblock In {\em Advances in Information Retrieval: 43rd European Conference
  on IR Research, ECIR 2021, Virtual Event, March 28--April 1, 2021,
  Proceedings, Part I 43}, pages 238--253. Springer, 2021.

\bibitem{glater2023answerpos}
Rafael Glater and Rodrygo L.~T. Santos.
\newblock On answer position bias in transformers for question answering.
\newblock In {\em Proceedings of the 46th International ACM SIGIR Conference on
  Research and Development in Information Retrieval}, SIGIR '23, page
  2215–2219, New York, NY, USA, 2023. Association for Computing Machinery.

\bibitem{peysakhovich2023attention}
Alexander Peysakhovich and Adam Lerer.
\newblock Attention sorting combats recency bias in long context language
  models.
\newblock {\em arXiv preprint arXiv:2310.01427}, 2023.

\bibitem{jiang2023longllmlingua}
Huiqiang Jiang, Qianhui Wu, Xufang Luo, Dongsheng Li, Chin-Yew Lin, Yuqing
  Yang, and Lili Qiu.
\newblock Longllmlingua: Accelerating and enhancing llms in long context
  scenarios via prompt compression.
\newblock {\em arXiv preprint arXiv:2310.06839}, 2023.

\bibitem{an2024make}
Shengnan An, Zexiong Ma, Zeqi Lin, Nanning Zheng, and Jian-Guang Lou.
\newblock Make your llm fully utilize the context.
\newblock {\em arXiv preprint arXiv:2404.16811}, 2024.

\bibitem{wang2022language}
Thomas Wang, Adam Roberts, Daniel Hesslow, Teven Le~Scao, Hyung~Won Chung,
  Iz~Beltagy, Julien Launay, and Colin Raffel.
\newblock What language model architecture and pretraining objective works best
  for zero-shot generalization?
\newblock In {\em ICML}, 2022.

\bibitem{tay2022unifying}
Yi~Tay, Mostafa Dehghani, Vinh~Q Tran, Xavier Garcia, Dara Bahri, Tal Schuster,
  Huaixiu~Steven Zheng, Neil Houlsby, and Donald Metzler.
\newblock Unifying language learning paradigms.
\newblock {\em arXiv preprint arXiv:2205.05131}, 2022.

\bibitem{medquad}
Asma {Ben Abacha} and Dina Demner{-}Fushman.
\newblock A question-entailment approach to question answering.
\newblock {\em {BMC} Bioinform.}, 20(1):511:1--511:23, 2019.

\bibitem{kwiatkowski2019natural}
Tom Kwiatkowski, Jennimaria Palomaki, Olivia Redfield, Michael Collins, Ankur
  Parikh, Chris Alberti, Danielle Epstein, Illia Polosukhin, Jacob Devlin,
  Kenton Lee, et~al.
\newblock Natural questions: a benchmark for question answering research.
\newblock {\em Transactions of the Association for Computational Linguistics},
  7:453--466, 2019.

\bibitem{devlin2018bert}
Jacob Devlin, Ming-Wei Chang, Kenton Lee, and Kristina Toutanova.
\newblock Bert: Pre-training of deep bidirectional transformers for language
  understanding.
\newblock In {\em ACL}, 2018.

\bibitem{hinton2012improving}
Geoffrey~E Hinton, Nitish Srivastava, Alex Krizhevsky, Ilya Sutskever, and
  Ruslan~R Salakhutdinov.
\newblock Improving neural networks by preventing co-adaptation of feature
  detectors.
\newblock {\em arXiv preprint arXiv:1207.0580}, 2012.

\bibitem{tunstall2023zephyr}
Lewis Tunstall, Edward Beeching, Nathan Lambert, Nazneen Rajani, Kashif Rasul,
  Younes Belkada, Shengyi Huang, Leandro von Werra, Cl{\'e}mentine Fourrier,
  Nathan Habib, et~al.
\newblock Zephyr: Direct distillation of lm alignment.
\newblock {\em arXiv preprint arXiv:2310.16944}, 2023.

\bibitem{xia2023training}
Mengzhou Xia, Mikel Artetxe, Chunting Zhou, Xi~Victoria Lin, Ramakanth
  Pasunuru, Danqi Chen, Luke Zettlemoyer, and Ves Stoyanov.
\newblock Training trajectories of language models across scales.
\newblock In {\em ACL}, 2023.

\bibitem{tay2022scale}
Yi~Tay, Mostafa Dehghani, Jinfeng Rao, William Fedus, Samira Abnar, Hyung~Won
  Chung, Sharan Narang, Dani Yogatama, Ashish Vaswani, and Donald Metzler.
\newblock Scale efficiently: Insights from pretraining and finetuning
  transformers.
\newblock In {\em ICLR}, 2022.

\bibitem{liu-etal-2023-g}
Yang Liu, Dan Iter, Yichong Xu, Shuohang Wang, Ruochen Xu, and Chenguang Zhu.
\newblock {G}-eval: {NLG} evaluation using gpt-4 with better human alignment.
\newblock In {\em EMNLP}, 2023.

\bibitem{wolf2019huggingface}
Thomas Wolf, Lysandre Debut, Victor Sanh, Julien Chaumond, Clement Delangue,
  Anthony Moi, Pierric Cistac, Tim Rault, R{\'e}mi Louf, Morgan Funtowicz,
  et~al.
\newblock Huggingface's transformers: State-of-the-art natural language
  processing.
\newblock {\em arXiv preprint arXiv:1910.03771}, 2019.

\bibitem{rasley2020deepspeed}
Jeff Rasley, Samyam Rajbhandari, Olatunji Ruwase, and Yuxiong He.
\newblock Deepspeed: System optimizations enable training deep learning models
  with over 100 billion parameters.
\newblock In {\em Proceedings of the 26th ACM SIGKDD International Conference
  on Knowledge Discovery \& Data Mining}, pages 3505--3506, 2020.

\bibitem{kang2019decoupling}
Bingyi Kang, Saining Xie, Marcus Rohrbach, Zhicheng Yan, Albert Gordo, Jiashi
  Feng, and Yannis Kalantidis.
\newblock Decoupling representation and classifier for long-tailed recognition.
\newblock {\em arXiv preprint arXiv:1910.09217}, 2019.

\bibitem{zhou2020bbn}
Boyan Zhou, Quan Cui, Xiu-Shen Wei, and Zhao-Min Chen.
\newblock Bbn: Bilateral-branch network with cumulative learning for
  long-tailed visual recognition.
\newblock In {\em CVPR}, pages 9719--9728, 2020.

\end{thebibliography}

\appendix
\setcounter{figure}{0}
\setcounter{table}{0}

\renewcommand{\thetable}{\Alph{table}}
\renewcommand{\thefigure}{\Alph{figure}}

\section{Broader Impact}
The general negative societal impacts of LLM will be applied to our work. 
Since our work focuses on adapting LLM to new information, the insight from our work might be used to adapt LLM to memorize private information. But, we think the issue is not specific to our work, but a general problem in LLM. To mitigate such issues, it is important to limit the use of private information to train LLM. 

\section{Details of dataset}

\subsection{Additional Details}\label{sec:dataset_detail}
\begin{table}[h]
    \centering
        \caption{Number of documents and QA pairs of used datasets. }  
    \begin{tabular}{c|c|c}
        \toprule
         Dataset    & Documents      & QA pairs\\
        \midrule
       \fontchange{pcr}{bioS} &3000& 27000\\
       \fontchange{pcr}{Wiki2023+} &5911& 10924\\
       \fontchange{pcr}{Film (Wiki2023+)}&2385&7398 \\
        \fontchange{pcr}{MedQuAD} &16407& 42687\\
       
        \bottomrule
    \end{tabular}
    \label{tab:neurips_dataset}
\end{table}

\textbf{Stats of dataset.}
Table~\ref{tab:neurips_dataset} summarizes the number of documents and QA pairs per dataset. 
\begin{figure}[h]
    \centering
\includegraphics[width=\textwidth]{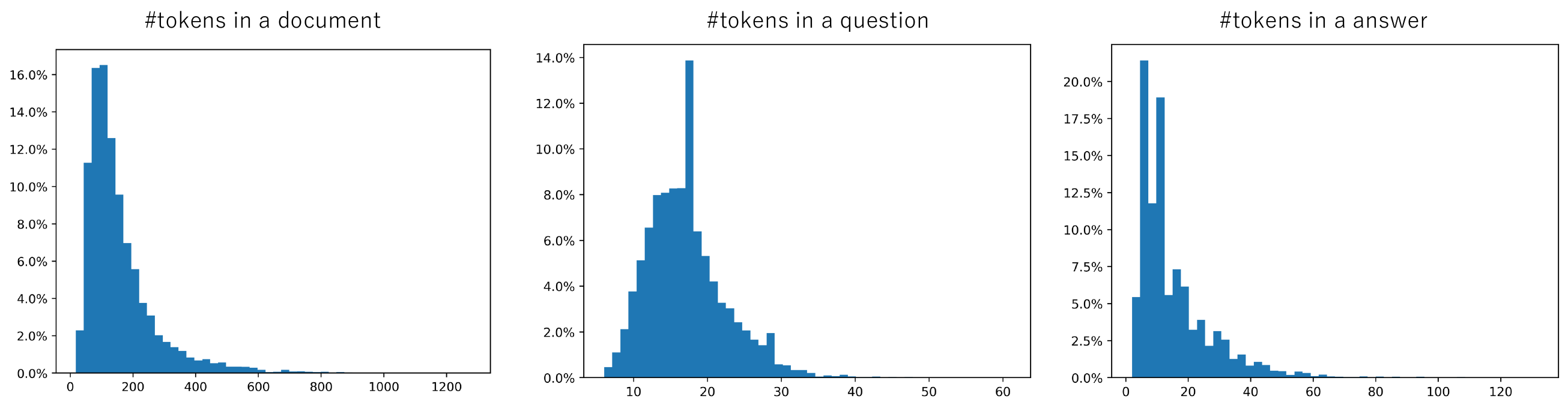}
    \caption{Histograms of the number of tokens in a document, question, and answer (from left to right). }
    \label{fig:stats_dataset}
\end{figure}
Fig.~\ref{fig:stats_dataset} illustrates the number of tokens per document, question, and answer sentence. Most learned documents are short, and questions and answers are generally concise. 
\begin{figure}[h!]
    \centering
\includegraphics[width=0.4\textwidth]{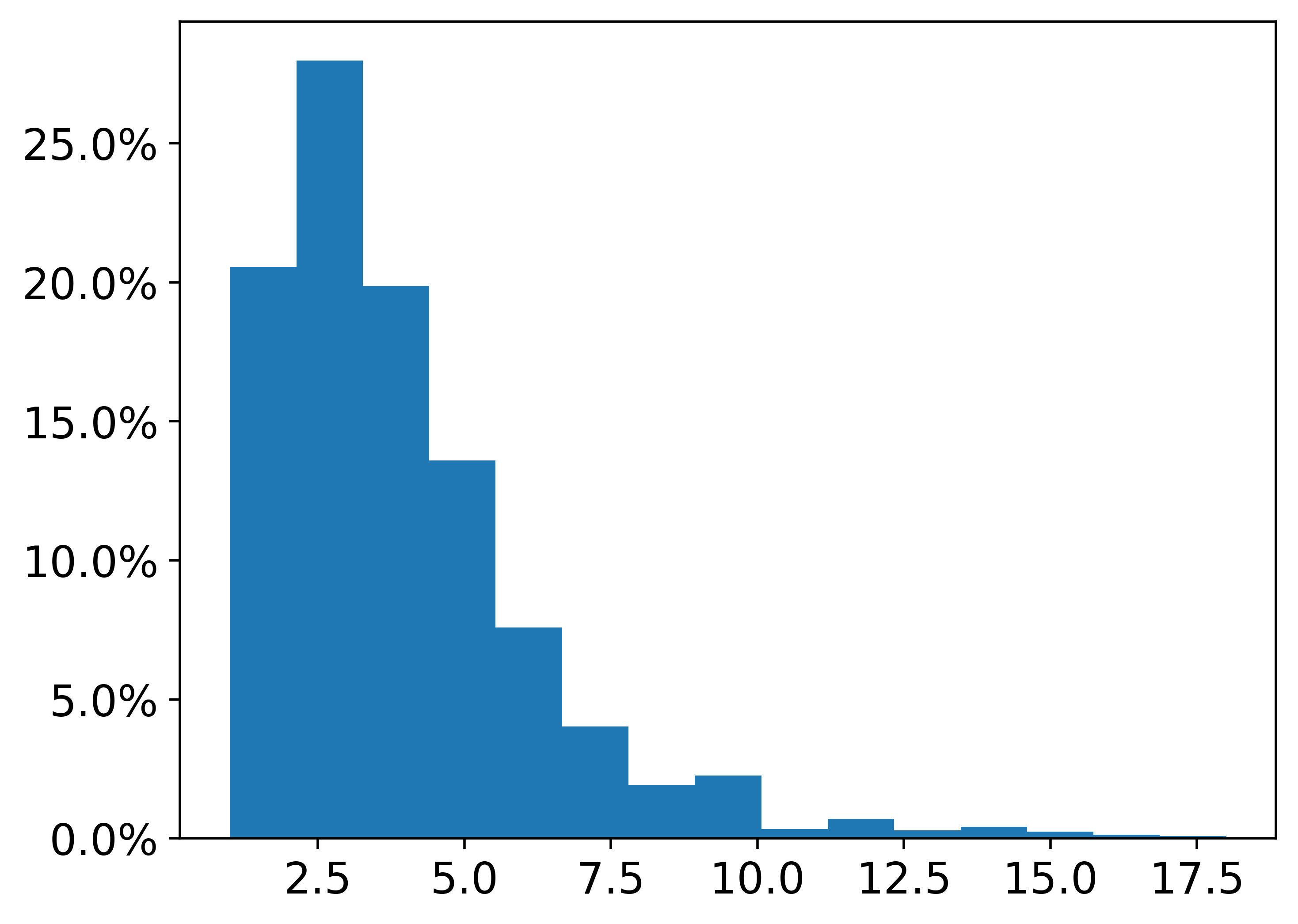}
    \caption{Histogram of the number of sentences per document (\fontchange{pcr}{Wiki2023+}).}
    \label{fig:num_sentence_dist}
\end{figure}
Fig.~\ref{fig:num_sentence_dist} illustrates the distribution of the number of sentences per document. The distribution is skewed as expected from the number of tokens (Fig.~\ref{fig:stats_dataset}).

\textbf{Investigation on the data leak.} 
Though \fontchange{pcr}{Wiki2023+} is collected from the articles published in 2023, there is a potential data leak to LLama-2 models. We feed the collected questions into LLama-2 7B model and measure the accuracy by asking Chat-GPT as described in Sec.~\ref{sec:details_exp}, obtaining an accuracy of 1.1\%. Note that the accuracy of Zephyr is 7.7\%. As indicated by \cite{jiang2024instruction}, we conclude that the data leak to the LLama-2 model is not significant.

\begin{figure}[ht]
    \centering
\includegraphics[width=0.9\textwidth]{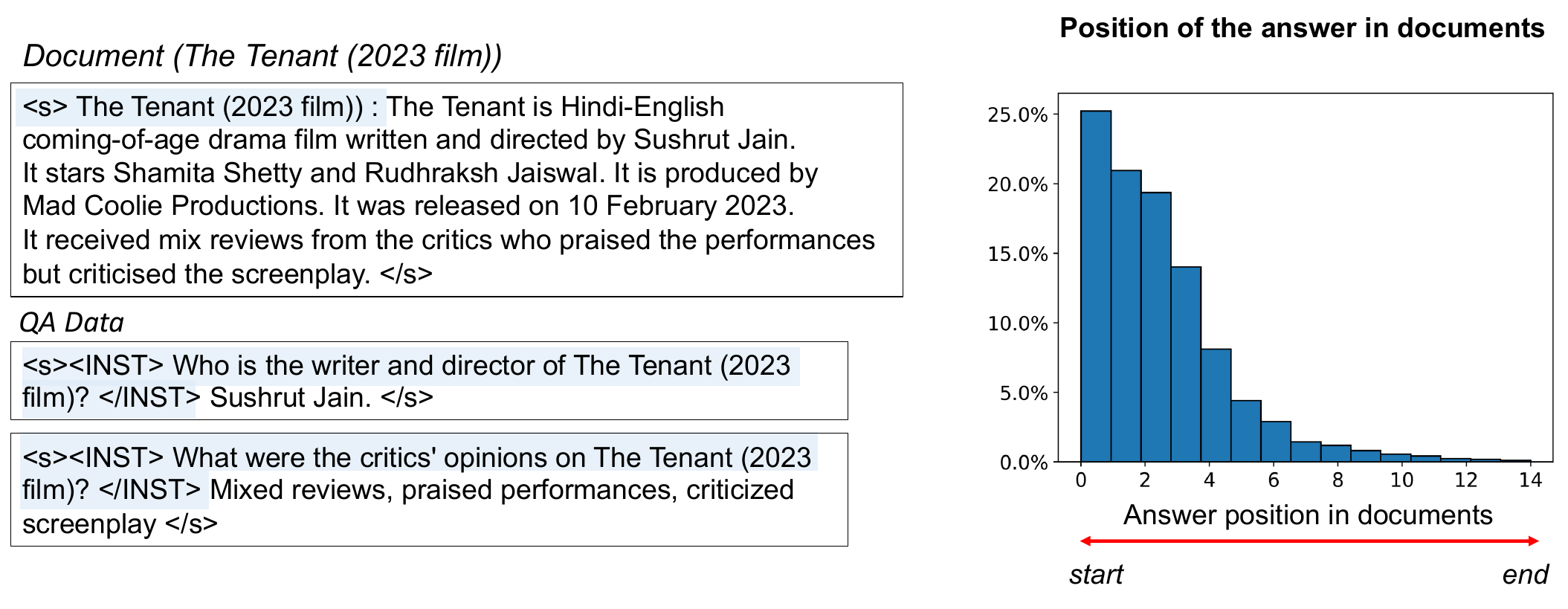}
    \caption{Left: Example of a document and QA pairs generated from the document. "<INST>" and "</INST>" are the tags used for LLama-2 Chat model. Right: The distribution of the position of answers in this dataset. The distribution is skewed towards the first sentence. }
    \label{fig:wiki2023}
\end{figure}

\textbf{Details of MedQuAD.} MedQuAD~\cite{medquad} includes 16407 medical question-answer pairs created from twelve NIH websites. The answer is often long and provides detailed information about the subject asked in the question. Then, we regard the answer as the knowledge documents and create new QA pairs for evaluation using ChatGPT, as shown in Sec.\ref{sec:qa_create}. During training, we utilize both original and our created QA pairs. Unlike \fontchange{pcr}{bioS} and \fontchange{pcr}{Wiki2023+}, this dataset is not annotated with the answer position. The evaluation is conducted without considering the position of the answers. 

\begin{figure}[t]
    \centering
\includegraphics[width=\textwidth]{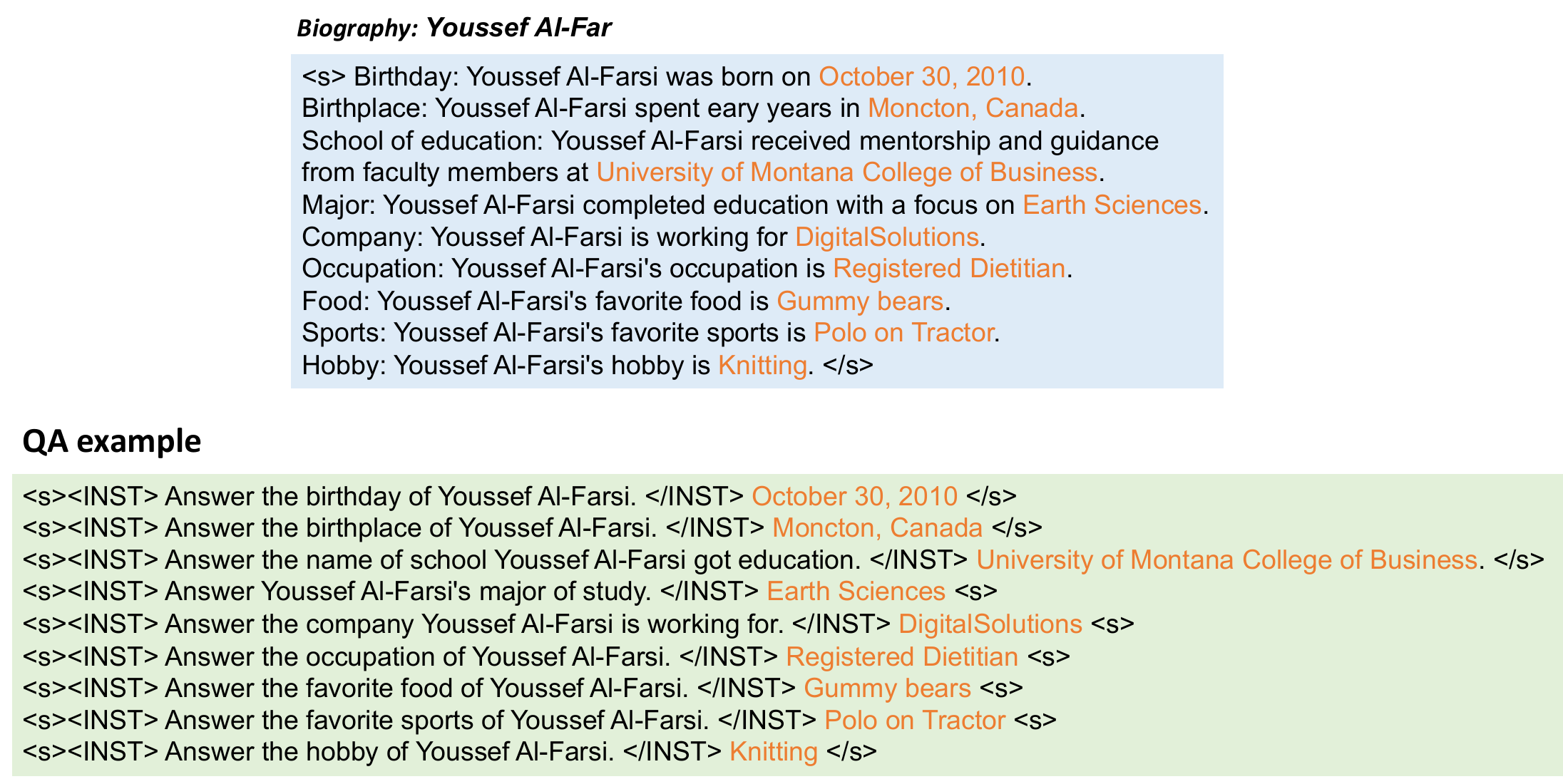}
    \caption{Example of \fontchange{pcr}{bioS}, and corresponding questions. Phrases unique to each person are highlighted with color. These parts are asked during evaluation.}
    \label{fig:synth_example}
\end{figure}
\textbf{Examples of documents.} 
Left of Fig.~\ref{fig:wiki2023} and Fig.~\ref{fig:synth_example} shows the example of \fontchange{pcr}{Wiki2023+} and \fontchange{pcr}{bioS} respectively. For Fig.~\ref{fig:wiki2023}, we highlight the tokens used as prompts to generate document contents or answers. 

\textbf{Question distributions.} 
The right of Fig.~\ref{fig:wiki2023} illustrates the histogram of the answer position in documents for the \textit{film} test split, revealing that the distribution of the answer position is skewed towards the beginning of the documents. This skewness arises from the presence of the first sentence in all documents, while some lack a second or subsequent one. Therefore, even if humans annotate the QA pairs for the entire document, not in a sentence-by-sentence way, this skewness can be present.

\begin{figure}[ht]
    \centering
\includegraphics[width=\textwidth]{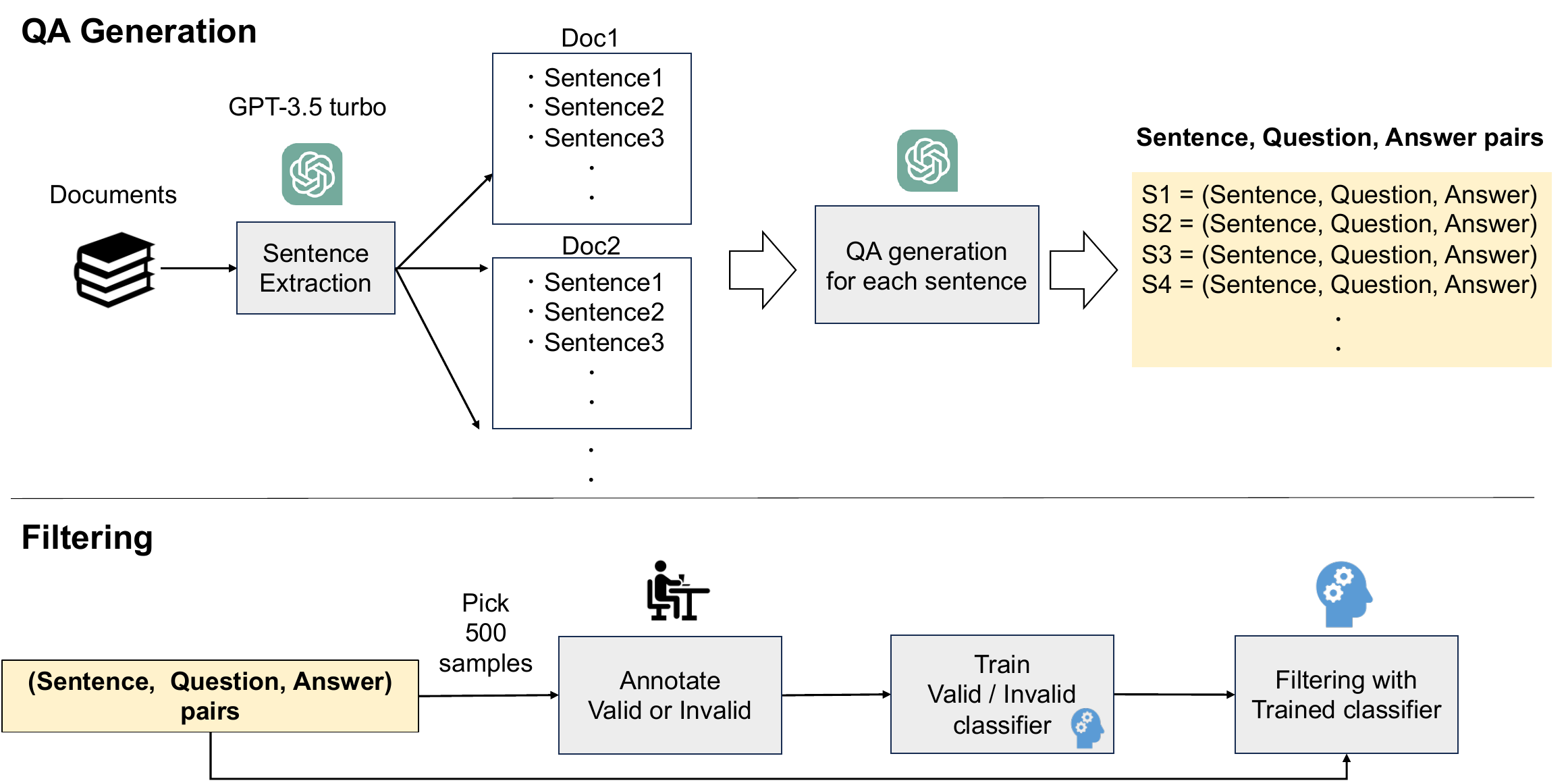}
    \caption{Procedure to create QA data for \fontchange{pcr}{Wiki2023+}.}
    \label{fig:qa_create_process}
\end{figure}
\subsection{Procedure to create QA pairs}\label{sec:qa_create}
Fig.~\ref{fig:qa_create_process} illustrates the overview of the QA dataset creation for \fontchange{pcr}{Wiki2023+}. 

\textbf{Sentence extraction.} First, we extract sentences from the documents. We find that splitting sentences using Chat-GPT works better than rule-based splitting or using the NLTK tool. Since the LLM can hallucinate some sentences, we compute the similarity between the documents before and after sentence extraction and filter documents if the similarity is smaller than a threshold. 

\textbf{QA generation.} We feed the sentence Chat-GPT and instruct the model to generate QA pairs. For the prompt sentence, we follow~\cite{jiang2024instruction}. 
The generated question needs to satisfy the following conditions; (i) it asks about the subject of the sentence, (ii) the answer can be inferred from the input sentence, and (iii) the answer cannot be the name of the subject. 
We find that the generated QA pairs include many samples that do not satisfy these conditions, approximately 20-30\% violate the conditions. 

\textbf{Annotate a small number of samples.}
Considering the noise in the QA pairs, we need to pick valid QA pairs. 
Since annotating all QA pairs takes tremendous cost, we choose to annotate a small proportion of samples and consider filtering the dataset by a classifier to judge the quality of the QA. 
Then, we randomly pick 500 samples from the QA pool and annotate the quality of QA data, \ie, valid or invalid under the conditions described above. 

\textbf{Filtering with a classifier.}
We train a sentence BERT-based~\cite{devlin2018bert} classifier by the 500 annotated samples. The input consists of the triplet of question, answer, and input sentence. Using this classifier, we pick QA samples with the top 60\% validity score. For the quality assessment, we further annotate 500 samples and confirm that 95\% of QA pairs satisfy the three conditions. 
Stats of samples that pass the filtering process are described in Table~\ref{tab:neurips_dataset}.

\section{Experimental Details}\label{sec:details_exp}

\textbf{Hyper-parameters.} In D-AR, the ratio to replace a token with a random one is set as 0.2 in all experiments. For Attn Drop, we set the dropout ratio as 0.5 in the experiments on \fontchange{pcr}{bioS} and 0.2 on \fontchange{pcr}{Wiki2023+}. For MedQuAD, we train all models for 6000 iterations considering the size of the dataset.

\begin{figure}[t]
    \centering
\includegraphics[width=0.8\textwidth]{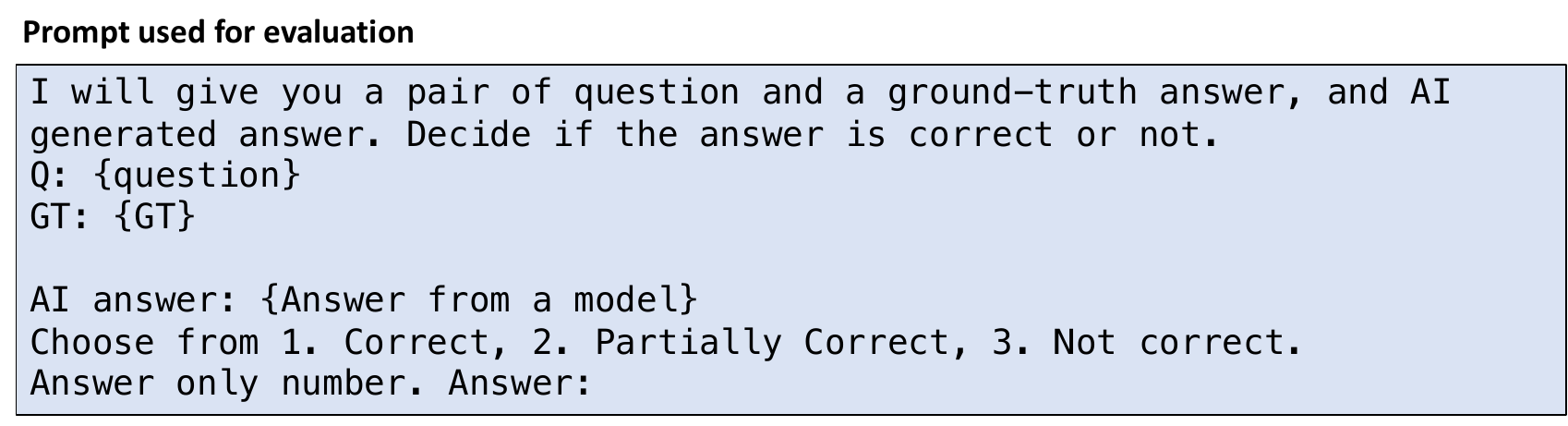}
    \caption{Example of a prompt used for evaluation.}
    \label{fig:eval_prompt}
\end{figure}

\textbf{Evaluation with ChatGPT.} 
Fig.~\ref{fig:eval_prompt} illustrates the prompt given to ChatGPT to evaluate the accuracy of the predicted answer given question and ground-truth. We utilize the percentage of \textit{Correct} instances as accuracy. 

\textbf{Implementation.} 
We implement our codebase relying on huggingface models~\cite{wolf2019huggingface} and utilize ZeRO3 in DeepSpeed~\cite{rasley2020deepspeed} for computational efficiency. During training, we set the number of training tokens to 512. 

\textbf{Inference.} 
We set the temperature as 0.6, top-k as 50, and repetition penalty as 1.2 in huggingface's text generation function. 

\textbf{Computation.} 
We employ a server with 8 A100 GPUs with either 40G or 80G memories. Our training code on LLama-2 7B occupies approximately 18G for each GPU. 

\textbf{Statistical Significance.} Due to the limit of time and resources, all results except for Fig.~\ref{fig:summary_result} are obtained by a single run. Fig.~\ref{fig:summary_result} shows the results averaged over three runs and their standard deviation.

\section{Additional Experiments}\label{sec:appendix_exp}
\textbf{Which is important, denoising or adding noise?} The denoising auto-regressive model shows remarkable improvements over a vanilla model in our experiments. In the D-AR model, some input tokens are replaced with random ones, and the model learns to predict the next \textit{correct} tokens. We study if the improvement comes from denoising the noise-added tokens or predicting next-tokens given randomly perturbed observations. Specifically, we turn off computing loss on the positions where the tokens are replaced with random ones, thereby ablating the denoising role. We evaluate it in the setting of Table~\ref{tab:position_wiki} and get an average of 29.3 / 54.8 (EM / F1). Compared to the vanilla D-AR model's performance (30.4 / 55.3), we see a small decrease, yet still surpassing the AR model by a large margin. We conclude that the performance gain comes largely from adding noise to the training tokens. 

\textbf{The effect of balanced sampling for QA is limited.} As shown in Fig.~\ref{fig:wiki2023}, the distribution of the answer position in QA data is highly skewed. A potential remedy to the imbalanced data is applying balanced sampling as done in long-tailed class recognition\cite{kang2019decoupling, zhou2020bbn}. Then, according to the number of samples in Fig.~\ref{fig:wiki2023}, we re-sample the QA samples in tail positions, resulting in averaged EM and F1 are 15.6 and 36.1 respectively (14.9 and 35.7 in the vanilla model). In summary, simply balancing QA samples by their positional distribution does not significantly improve the performance. 


\begin{table*}[h]
\caption{Detailed results on \fontchange{pcr}{bioS}.}
\scalebox{0.9}{
\centering
\begin{tabular}{c|c|c|c|c|c|c|c|c|c||c}
\toprule
Method&Birthday & Birthplace & School & Major & Company & Job & Food & Sports & Hobby & Avg     \\\hline

AR& 98.4&99.0 & 89.5 & 90.3 & 88.1 & 81.7 & 94.2 & 86.1 & 78.7 & 89.6 \\
D-AR & 98.4 & 99.2 & 97.8 & 99.2 & 97.0 & 96.6 & 96.0 & 97.6 & 96.2 & 97.6\\ 
Shuffle & 75.8 & 87.9 & 92.5 & 81.2 & 91.5 &  95.8 & 91.5 & 93.5& 91.9 & 89.0 \\ 
Attn Drop & 99.4 & 99.2 & 98.2 & 93.8 & 93.4 & 94.4 & 93.0 & 93.0 & 92.8 & 95.2 \\
\bottomrule
\end{tabular}}
\label{tab:bios_detail}
\end{table*}
\textbf{Detailed results on} \fontchange{pcr}{bioS}. Table~\ref{tab:bios_detail} describes the detailed results on \fontchange{pcr}{bioS}, which are omitted from the main draft due to the limit of space. From left to right, the contents are arranged in order from the first to the last. The result shows a trend similar to Fig.~\ref{fig:summary_result}, \ie, the AR model drops the performance on the second and third groups.

\begin{figure}[ht]
    \centering
\includegraphics[width=0.45\textwidth]{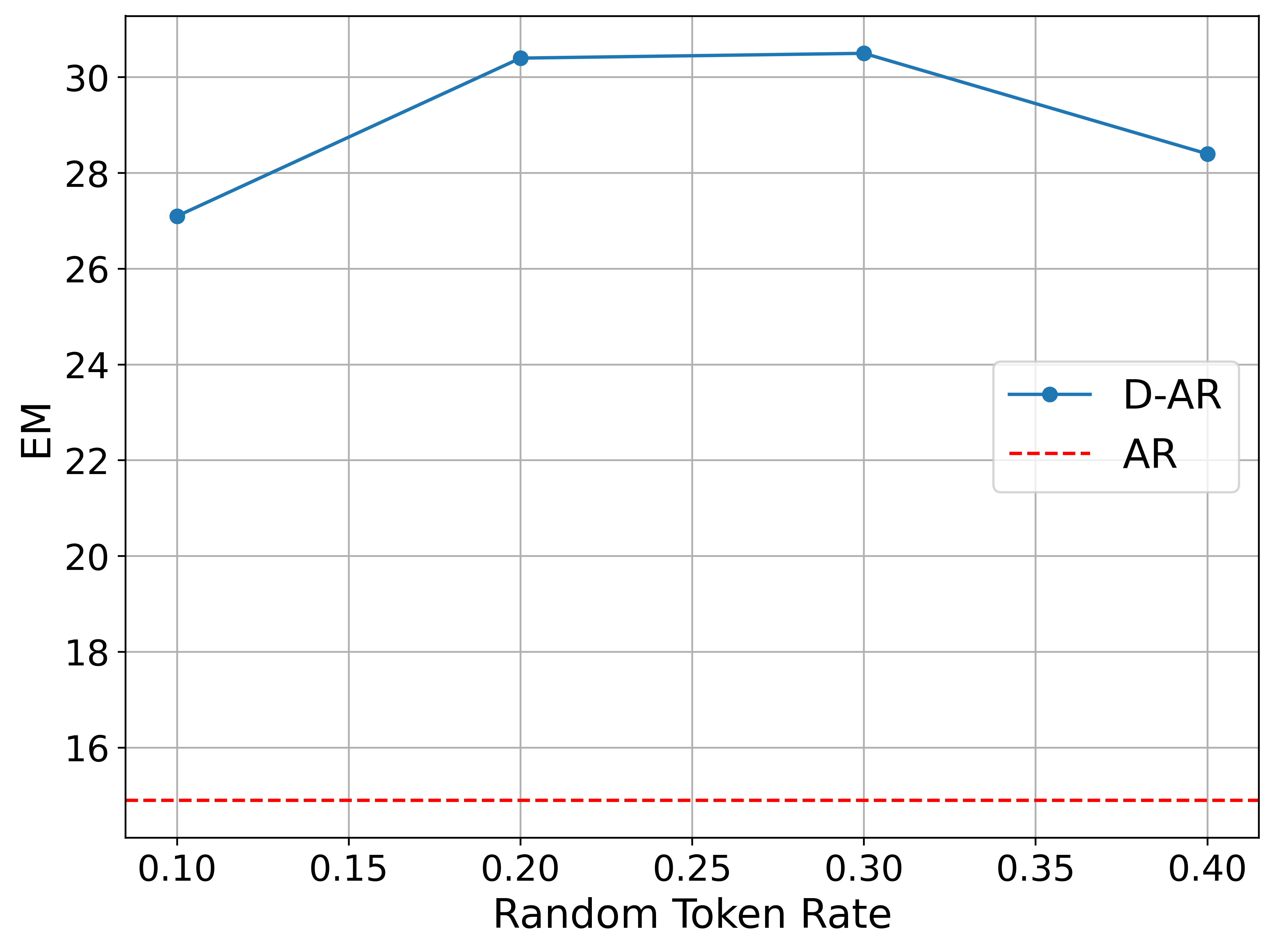}
    \caption{Ratio to add noise to input tokens in D-AR. EM (y-axis) and the position of the answer in the document (x-axis) in \fontchange{pcr}{Wiki2023+}.}
    \label{fig:mask_rate_wiki}
\end{figure}

\textbf{Sensitivity to hyper-parameter. }
Fig.~\ref{fig:mask_rate_wiki} describes the sensitivity to the ratio to replace input tokens with random ones in D-AR. Each plot shows the results of EM averaged over locations. Considering the performance of the AR model, adding small noise into the input sequences significantly improves the performance. 

\begin{table*}[h]
\centering
\newcolumntype{Y}{>{\raggedright\arraybackslash}X}

\caption{D-AR on QA data does not improve performance. We see significant degrade by applying D-AR objective to QA data. }
\scalebox{0.8}{
\begin{tabular}{c|cccccc|c}
\toprule
\multirow{2}{*}{Denoising}&\multicolumn{6}{c|}{$\longleftarrow$ start------------------------------------------------------------end$\longrightarrow$}&\multirow{2}{*}{Avg.}\\
&EM$_1$ / F1$_1$  &EM$_2$ / F1$_2$ &EM$_3$ / F1$_3$&EM$_4$ / F1$_4$&EM$_5$ / F1$_5$&EM$_6$ / F1$_6$&     \\\hline
Document&\textbf{60.1} / \textbf{73.7}& \textbf{26.9} / \textbf{53.1}& \textbf{23.4} / \textbf{52.9}& \textbf{26.0} / \textbf{51.7} & \textbf{24.8} / \textbf{52.2}& \textbf{21.3} / \textbf{48.2}&\textbf{30.4} / \textbf{55.3}\\ 
Document + QA& 52.6 / 66.0 & 20.7 / 43.0 & 15.9 / 43.2 & 21.9 / 48.8 & 19.4 / 51.2 & 15.7 / 45.7 & 24.4 / 49.7\\
\bottomrule
\end{tabular}}
\label{tab:dar_qa}
\end{table*}

\textbf{Does D-AR on QA data improve performance?} 
In Table~\ref{tab:dar_qa}, we examine the effectiveness of D-AR on the QA dataset. Specifically, we apply the random token replacement to both QA and document data. However, we see significant decreases in the performance. 

\begin{table*}[h]
\centering
\newcolumntype{Y}{>{\raggedright\arraybackslash}X}

\caption{Comparison by models in \fontchange{pcr}{Wiki2023+}. Zephyr's performance is comparable to LLama-70B model.}
\scalebox{0.8}{
\begin{tabular}{cc|cccccc|c}
\toprule
\multirow{2}{*}{Model Size}&\multirow{2}{*}{Method}&\multicolumn{6}{c|}{$\longleftarrow$ start------------------------------------------------------------end$\longrightarrow$}&\multirow{2}{*}{Avg.}\\
&&EM$_1$ / F1$_1$  &EM$_2$ / F1$_2$ &EM$_3$ / F1$_3$&EM$_4$ / F1$_4$&EM$_5$ / F1$_5$&EM$_6$ / F1$_6$&     \\\hline
\multirow{2}{*}{7B}&AR&40.9 / 54.0 & 6.3 / 20.5&8.1 / 29.8&11.7 / 35.7&11.6 / 37.8 &10.7 / 36.4&14.9 / 35.7\\
&D-AR&\textbf{60.1} / \textbf{73.7}& \textbf{26.9} / \textbf{53.1}& \textbf{23.4} / \textbf{52.9}& \textbf{26.0} / \textbf{51.7} & \textbf{24.8} / \textbf{52.2}& \textbf{21.3} / \textbf{48.2}&\textbf{30.4} / \textbf{55.3}\\\hline
\multirow{2}{*}{13B}&AR&58.1 / 69.3 & 8.7 / 28.3 & 	18.8 / 40.2 & 20.2 / 42.3 & 11.6 / 39.2 & 14.7 / 39.3 & 22.0 / 43.1 \\
&D-AR&\textbf{67.6} / \textbf{84.1} & \textbf{34.4} / \textbf{64.4}& \textbf{32.8} / \textbf{64.0} & \textbf{30.5} / \textbf{58.6}& \textbf{30.2} / \textbf{59.0} & \textbf{22.8} / \textbf{52.2} & \textbf{36.4} / \textbf{63.7}\\\hline
\multirow{2}{*}{70B}&AR&65.3 / 78.9& 27.2 / 48.9& 24.4 / 46.2& 27.8 / 50.9& 22.5 / 50.9& 22.8 / 48.1& 31.7 / 54.0\\ 
&D-AR & \textbf{70.8} / \textbf{85.8} & \textbf{48.8} / \textbf{68.9} & \textbf{43.8} / \textbf{70.7} & \textbf{39.5} / \textbf{64.8} & \textbf{38.0} / \textbf{66.7} & \textbf{36.0} / \textbf{60.7} & \textbf{46.2} / \textbf{69.6}\\\hline
\multirow{2}{*}{Zephyr} &AR& 64.8 / 79.7 & 18.6 / 45.7 & 25.9 / 52.3 & 32.2 / 54.0 & 21.7 / 52.5 & 22.3 / 48.8 & 30.9 / 55.5\\
&D-AR& \textbf{73.6} / \textbf{88.0} & \textbf{46.7} / \textbf{70.3} & \textbf{42.9} / \textbf{71.9} & \textbf{39.4} / \textbf{66.9} & \textbf{36.4} / \textbf{66.5} & \textbf{31.4} / \textbf{57.6} & \textbf{45.1} / \textbf{70.2} \\
\bottomrule
\end{tabular}}
\label{tab:model_compare_wiki}
\end{table*}
\textbf{Results of Zephyr.} 
Table~\ref{tab:model_compare_wiki} presents the results of Zephyr in \fontchange{pcr}{Wiki2023+}. Overall, the performance is comparable to the Llama-70B model. As we indicate in the main paper, due to the potential data leak, it is hard to conclude that Zephyr is a better knowledge learner. Even if so, the cause of the better performance is not clear. Further investigation is our future work.

\end{document}